\pdfoutput=1
\documentclass[11pt]{article}

\usepackage[preprint]{acl}
\usepackage{listings}

\usepackage{times}
\usepackage{latexsym}
\usepackage{multirow}
\usepackage[T1]{fontenc}
\usepackage{tabularx}
\usepackage{adjustbox}
\usepackage[utf8]{inputenc}

\usepackage{microtype}

\usepackage{inconsolata}

\usepackage{graphicx}
\usepackage{algorithm}
\usepackage{algpseudocode}

\title{Accelerating Large Language Model Pretraining via LFR Pedagogy: \underline{L}earn, \underline{F}ocus, and \underline{R}eview }

\author{Neha Prakriya \\
  UCLA \\
  \texttt{nehaprakriya@cs.ucla.edu} \\\And
  Jui-Nan Yen \\
  UCLA \\
  \texttt{juinanyen@cs.ucla.edu} \\\AND
  Cho-Jui Hsieh \\
  Google and UCLA \\
  \texttt{chohsieh@cs.ucla.edu}\\\And
  Jason Cong \\
  UCLA \\
  \texttt{cong@cs.ucla.edu}\\}

\begin{document}
\maketitle
\begin{abstract}

Traditional Large Language Model (LLM) pretraining relies on autoregressive language modeling with randomly sampled data from web-scale datasets. Inspired by human learning techniques like spaced repetition, we hypothesize that random sampling leads to high training costs, lower-quality models, and significant data forgetting. To address these inefficiencies, we propose the Learn-Focus-Review (LFR) paradigm—a dynamic training approach that adapts to the model's learning progress. LFR tracks the model’s learning performance across data blocks (sequences of tokens) and prioritizes revisiting challenging regions of the dataset that are more prone to being forgotten, enabling better retention and more efficient learning. Using the LFR paradigm, we pretrained Llama and GPT models on the SlimPajama and OpenWebText datasets, respectively. These models were evaluated on downstream tasks across various domains, including question answering, problem-solving, commonsense reasoning, language modeling, and translation. Compared to baseline models trained on the full datasets, LFR consistently achieved \textbf{lower perplexity and higher accuracy}, while using only 5\%–19\% of the training tokens. Furthermore, LFR matched the performance of industry-standard Pythia models with up to 2$\times$ the parameter count, using just 3.2\% of the training tokens, demonstrating its effectiveness and efficiency.

\end{abstract}

\section{Introduction}
LLMs have achieved remarkable success in understanding and generating human language. This success is driven by the ever-increasing model parameter sizes which require web-scale training datasets like SlimPajama~\cite{cerebras2023slimpajama}, OpenWebText~\cite{Radford2019LanguageMA, openwebtext}, CommonCrawl~\cite{crawl, raffel2023exploring}, and Pile~\cite{gao2020pile}, leading to unsustainable training costs.
% Between 2016 and 2023, model parameter sizes have skyrocketed by a factor of 410x every two years.
% Meanwhile, GPU memory has scaled at a much slower pace of 2x every 2 years~\cite{gholami2024ai}, increasing model training costs exponentially.
For example, pretraining the GPT-4 model~\cite{openai2024gpt4} was estimated to have cost around \$100M USD
over a period of 3-4 months using 25k NVIDIA A100 GPUs~\cite{gpt4-openai}.
As such, a key challenge for unlocking the next generation of language models is to
significantly reduce training costs while retaining or improving downstream task performance.

Popular LLMs, like the GPT series~\cite{Radford2019LanguageMA, NEURIPS2020_1457c0d6, openai2024gpt4}, Llama series~\cite{touvron2023llama, llama3},  Gemini~\cite{geminiteam2024geminifamilyhighlycapable}, Mistral~\cite{jiang2023mistral}, and others, were trained utilizing autoregressive/causal techniques, where the goal of training is to generate a plausible next token based on a randomly sampled input data block from the training corpus.
However, data scraped from the web is highly redundant and presents varying complexity. For example, Table~\ref{tab:1_examples_from_slimpajama} shows two randomly selected example texts from the SlimPajama~\cite{cerebras2023slimpajama} dataset. The SlimPajama dataset contains text from a variety of sources such as CommonCrawl, Github, ArXiv, Wikipedia, etc. If a human is asked to learn these two texts, they might spend more time learning the details and semantics of the code in the second block of text than the text in the first block. The second block contains more complex information with previously unseen code semantics, while the first block contains anecdotal information about a bakery, which can 
be understood faster. They might read the second example more times than the first. 
However, autoregressive language modeling does not mimic such behavior.

\begin{table}[h]
\centering
\begin{tabularx}{\columnwidth}{|X|}
\hline
"Bread made to order and delivered fresh to you from my micro-bakery. All ingredients are seasonal and locally sourced." Find Rise 'n' Shine Bakehouse regularly at our Emsworth Farmers' Market. Find Foccacia, bagels, baguettes, french toast, sourdough and much more! Which markets are we attending? \\
  \hline
  NAMESPACE\_PH0LY\_BEGIN(thread) class PH0LY CriticalSection { public: CriticalSection(); CriticalSection(); CriticalSection(const CriticalSection$\&$); CriticalSection$\&$ operator = (const CriticalSection$\&$ rhs); void Enter(); void Leave(); private: CRITICAL\_SECTION m\_cs; }; NAMESPACE\_PH0LY\_END \\
  \hline
\end{tabularx}
\caption{Examples of Text Blocks in the SlimPajama Dataset. The second block is considerably hard to learn.}
\label{tab:1_examples_from_slimpajama}
\end{table}

\begin{figure}
    \centering
\includegraphics[width=\columnwidth]{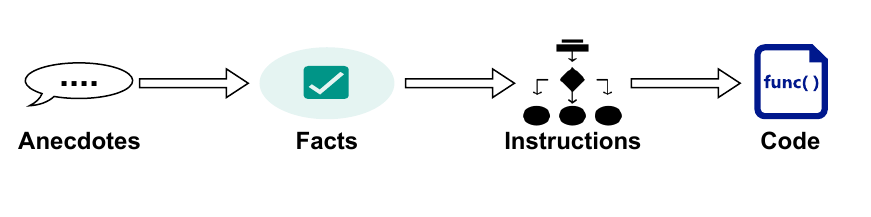}
    \caption{LLM training timeline}
    \label{llm-training-timeline}
\end{figure}

Inspired by these observations on different learning methods used by humans and LLMs, our work has the following contributions:
\begin{enumerate}
    \item Profile LLM pretraining to observe \textbf{multiple descent behaviour in 25-78\% of the training tokens}, which are forgotten multiple times during training.
    \item Develop a \textbf{Learn-Focus-Review (LFR) training pipeline} that uses perplexity to gauge the LLM's learning pace, focusing on complex data blocks while regularly reviewing all data blocks to prevent forgetting. 
    \item  Pretrain the Llama-2 models (120M-500M) using 4 NVIDIA A6000s and and the GPT-2 models (124M-1.5B) using 4 AMD MI210 and 4 MI100 GPUs from scratch on the SlimPajama and OpenWebText datasets and evaluate them on various downstream tasks. 
    % In fact, we believe that we are the first to train the GPT-2 models from scratch on a cluster of AMD GPUs, \textbf{demonstrating an alternative GPU platform for LLM training}. 
    \item LFR results in significantly \textbf{lower perplexity and higher accuracy} compared to baseline models trained on the full dataset using random sampling, achieving these improvements with \textbf{5.1$\times$ and 20$\times$ fewer training iterations for the Llama and GPT-2 models respectively}.
    \item Observe that LLMs first learn conversational and anecdotal data, before being able to retain factual, instructional, and coding language information in long-term memory. Furthermore, we compare our models with Pythia models of different scales trained on 30$\times$ more training tokens, and achieve comparable performance.
    \item Demonstrate that text importance varies with training time and model size, driving the need for dynamic data selection methods like LFR.

\end{enumerate}

% TODO add a paragraph to transition here
In the following sections, we examine prior works on efficient LLM pretraining before diving deeper into our proposed training strategies and design decisions. 

\section{Related Work}\label{prior_work}
Prior works on efficient pretraining of LLMs using data selection have primarily focused on using distance metrics and clustering techniques. \citet{tirumala2023d4} proposes D4, which deduplicates and selects cluster centers in the embedding space generated by pretrained models.
SemDeDup~\cite{abbas2023semdedup} prunes semantic duplicates using pretrained models. It can successfully prune 50\% of the training data with minimal performance loss. MiniPile~\cite{kaddour2023minipile, minipile-huggingface} uses the pretrained E5-Large~\cite{wang2024text} model to embed documents in the Pile dataset and clusters them to select a smaller pretraining corpus of $\sim$6GB. DSIR~\cite{xie2023data} proposes selecting subsets from large unlabeled datasets through importance resampling to match the distribution of the target dataset. However, considering the high cost of training, it is unsustainable to sample a new subset and pretrain the LLM from scratch for every new downstream task. \citet{lin2024not} proposes a high-performance selection methodology, but suffers from dependence on a reference model for selection.

While these works address the common redundancy issue present in LLM pretraining corpuses, they face three major drawbacks. First, they require pretrained models for calculating the distance metric on embeddings. This step must ensure that the embeddings found by the pretrained model are similar to those of the target model to ensure the transferability of sample importance. Second, \textit{static} clustering-based methods, which cluster data only at the start of training, fail to adapt to the diverse and dynamic nature of web-scale datasets. These methods overlook the fact that the difficulty of specific data points can vary across different training epochs, limiting their ability to effectively address the evolving learning needs of the model. This can lead to a loss in downstream task accuracy (see Section~\ref{eval}).

\citet{marion2023more} evaluates data selection by pretrained models used to obtain perplexity, Error L2-Norm (EL2N)~\cite{Paul2021DeepLO}, and memorization rankings~\cite{biderman2023emergent}. They find that perplexity is a good metric to characterize a model's understanding of a data sample and that retaining data samples with middle perplexity values can outperform training on the entire dataset. While this finding inspires our work, their method relies on pretrained model checkpoints to evaluate data sample importance. This limits the scalability. 
\begin{figure}
\centering\includegraphics[width=\columnwidth]{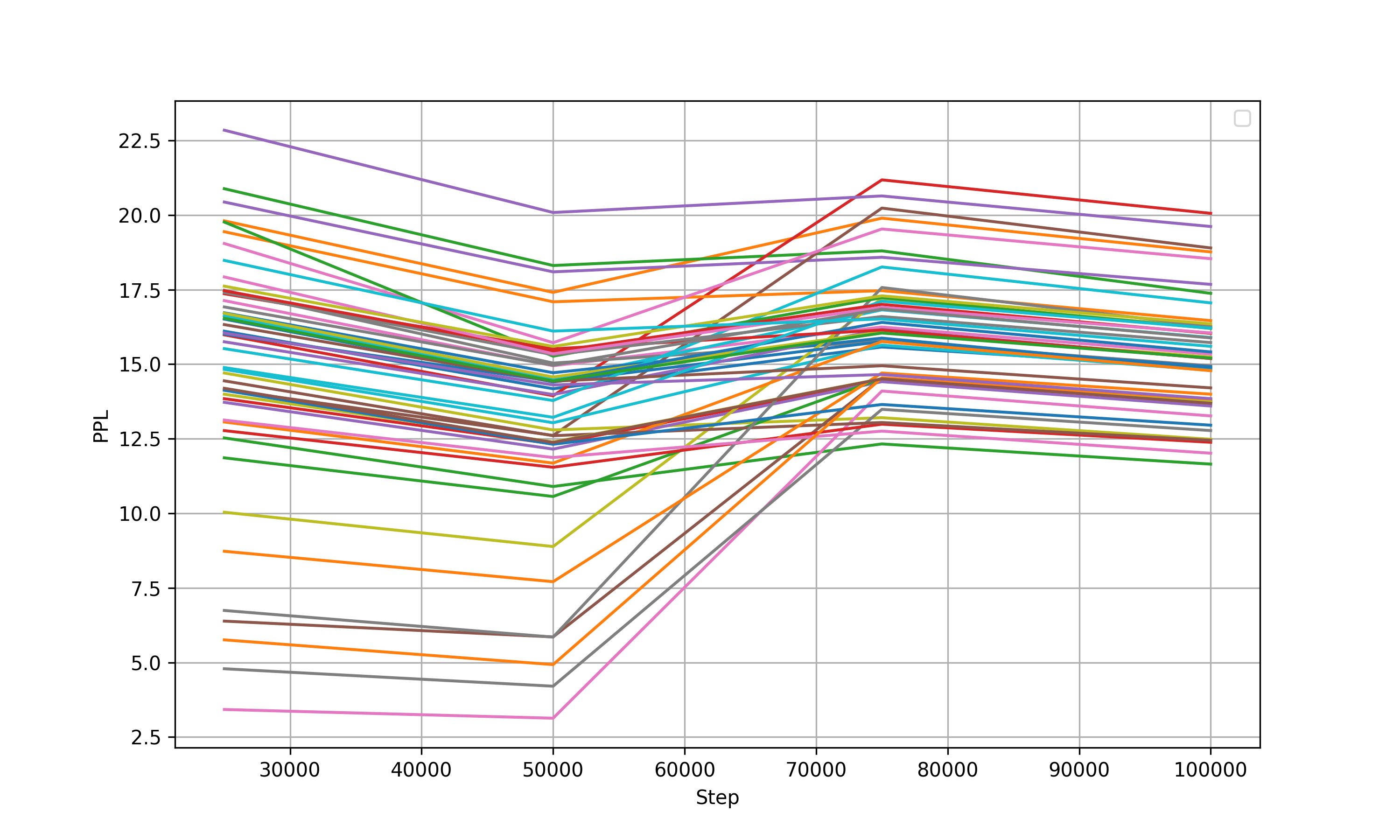}
    \caption{PPL trajectories of data samples from the SlimPajama dataset as processed by the Llama-300M model, focusing on a subset of 50 samples for clarity. Notably, 78.5\% of the samples exhibit this behavior, characterized by multiple descent patterns rather than a steady decline. This indicates that the model frequently forgets and relearns data during training, highlighting inefficiencies in traditional training dynamics}
    \label{llama-forgetting}
\end{figure}

\begin{figure}
    \centering \includegraphics[width=\columnwidth]{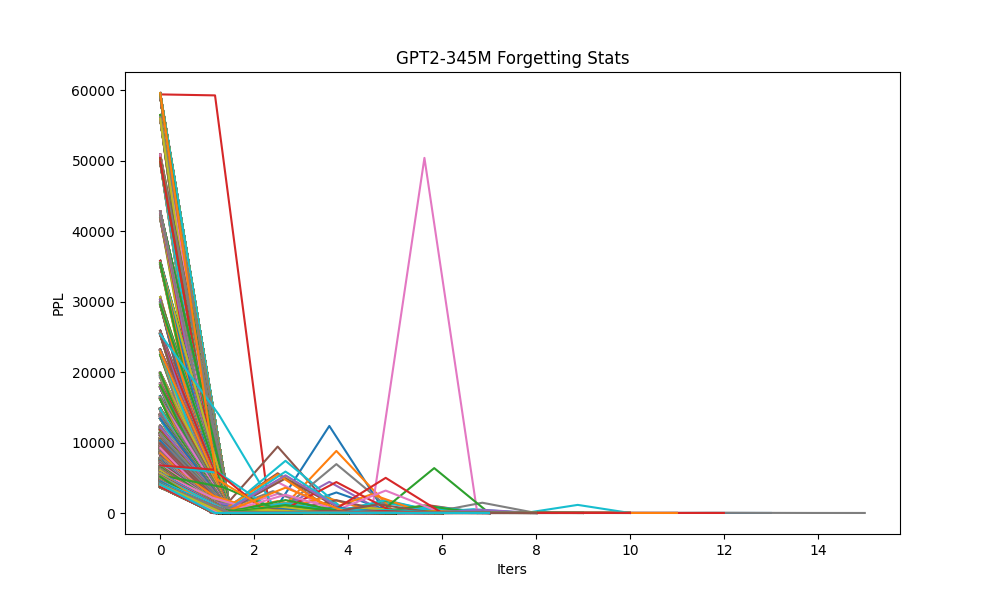}
    \caption{PPLs of data samples being forgotton by the GPT2-345M model. This multi-descent behavior is exhibited by 20\% of the data.}
    \label{fig:medium-forgetting}
\end{figure}
\section{Problem Formulation and Profiling} \label{method}
\subsection{LLM Pretraining Objective}
Given an LLM model parameterized by weights $\theta$ and a web-scale dataset $D$, we first tokenize all the documents in the dataset and obtain context-length sized sequences of tokens, called data blocks, $s_i$ such that the training corpus becomes $D = \{s_1, s_2, s_3, ... s_n\}$. For the SlimPajama and OpenWebText datasets used in this paper, the context length is 1024 tokens, with a total of 627B and 9B tokens, respectively. Given one such sequence of tokens or data block, $s_i = \{x_1, x_2, ... x_n\}$, the training objective is to autoregressively predict the next $M$ tokens:
\begin{equation}\label{eq:pretraining_objective}
    p_\theta (y \mid x) = \prod_{i=1}^{M} p_\theta (y_i \mid y_{1:i-1} , x).
\end{equation}
The batches of sequences ($s_i$) used as input to prompt the LLM (shown in Eq.~\ref{eq:pretraining_objective}) and backpropagate for each step are randomly sampled from the training corpus. 

\subsection{Observations from Training on Randomly Sampled Data}\label{obs}
In order to better understand the drawbacks of this traditional training technique, we profile the pretraining process for the 300M parameter Llama-2 model (single epoch training) and the 345M parameter GPT-2 model (multi-epoch training).
The training hyperparameters and model configurations are provided in Appendix~\ref{app_training_details}.
Similarly to~\citet{marion2023more}, we use \textit{perplexity} as a metric to monitor the training progress. 
Given a token sequence $s_i = \{x_1, x_2, ..., x_n\}$ from the dataset $D$, perplexity is computed as:

\begin{equation}\label{eq:PPL}
    PPL(s_i) = \exp \left( \frac{1}{|s_i|} \sum_{x_j \in s_i}NLL(x_j) \right),
\end{equation}
where $NLL(x_j)$ is the negative log likelihood of token $x_j$ computed as follows:
\begin{equation}\label{eq:nll}
    NLL(x_j) = -\log P(x_j \mid x_{<j};\theta).
\end{equation}
Using this metric, models exhibiting lower perplexities are considered better since this indicates a high probability of selecting text closest to the raw dataset.

% In the case of the Llama-300M model, we use the Chinchilla scaling law~\cite{hoffmann2022trainingcomputeoptimallargelanguage} to train the model for 100k steps with a total of 9.8B tokens from the SlimPajama dataset. 
The Llama models are trained with the SlimPajama dataset~\cite{cerebras2023slimpajama} for 100k steps where we measure and plot the $PPL$ associated with the training set. In the case of the GPT-2 model, we train it for 8 epochs such that each data point in the OpenWebText dataset has at least 8 PPL values captured at different training iterations. The observed PPL values associated with each data block are classified as one of the following:
\begin{enumerate}
    \item \textit{Learned}: recorded perplexities monotonically decrease. 
    \item \textit{Unlearned}: recorded perplexities monotonically increase.
    \item \textit{Forgotten}: recorded perplexities first increase and then decrease. Such an upward and downward trend may repeat any number of times during training. 
\end{enumerate}

Based on this classification, we observe that 78.5\% of the data blocks are forgotten at least once in the Llama-2 model (Figure~\ref{llama-forgetting}), compared to 25\% in the GPT-2 model (Figure~\ref{fig:medium-forgetting}). We hypothesize that more data blocks are frequently forgotten in the Llama model due to the higher complexity and challenge posed by the SlimPajama dataset, as opposed to the OpenWebText dataset. It is important to note that the SlimPajama dataset is an aggregation of seven datasets, including sources such as GitHub, Wikipedia, and CommonCrawl. In fact, of the data blocks that are forgotten, 82\% are forgotten multiple times during training, i.e., they display \textit{multiple descent behavior} (Figure~\ref{fig:medium-forgetting}). \citet{Xia2022TrainingTO} reported a double-descent behavior for the OPT models \cite{Zhang2022OPTOP}, and our above experiment further demonstrates that the ``forgetting'' can happen multiple times in LLM pretraining.

\section{LFR Training Methodology}\label{method}

Based on our profiling observations in Section~\ref{obs} we propose to replace traditional autoregressive language modeling methods with Spaced Repetition \cite{pnas-spacerep}. Spaced Repetition is an evidence-based learning method proven to improve information retention and learning pace in humans \cite{Smolen_2016}. In this technique, challenging pieces of information are reviewed more often, at regular intervals, and easier pieces of information are presented to the learner less often. Our algorithm is detailed in Algorithm \ref{alg:cap}. We pretrain our models with a combination of the following three steps:

\begin{enumerate}
    \item \textbf{Learn}: Initially, we present the model with the entire dataset and train on randomly selected data blocks for $p_1$ steps, as normally seen in the traditional approach (line 1 in Alg.~\ref{alg:cap}). $p_1$ can be configured by the user based on the available compute budget, model, and dataset. In single-epoch training (lines 3-7 in Alg.~\ref{alg:cap}), we measure the perplexities (PPLs) of all data samples in the training set and cluster the data embeddings (inputs to the model's last layer). 
    % Since the SlimPajama dataset used to train the Llama models consists of seven data sources, we set $k = 7$ for clustering. 
    For multi-epoch training (lines 8-11 in Alg~\ref{alg:cap}), we record the perplexities for all data blocks during the $p_1$ steps. Depending on the training style (single or multi-epoch), we either pass the clustered embeddings and PPL values or the PPL values observed during training to the next step. The following two phases can be repeated up to $reps$ times, depending on the available compute budget.
    \item \textbf{Focus}: We provide two variations of the Focus stage based on the number of training epochs. 
    \begin{enumerate}
    \item Single-epoch training: We discard $s_1$\% of the clusters (line 6 in Alg~\ref{alg:cap}). Within the retained clusters, we perform weighted sampling from sub-clusters, prioritizing regions of the retained clusters which the model finds most challenging (line 7 in Alg.~\ref{alg:cap}). Sub-clusters with higher $PPL$ are assigned greater sampling weights, enabling a hierarchical focus on the most critical regions. For instance, during Llama-2 training, GitHub code emerged as the most challenging cluster. Within this cluster, the Focus stage further emphasizes sampling from C++ code, which proved more difficult for the model, over Python code. In this phase of training, we restrict the weighted sampling of data points to this reduced subset for $p_2$ steps. $s_1$ and $p_2$ are user-controlled hyperparameters.
    \item Multi-epoch training: We discard $s_1$\% of the data blocks (line 10 in Alg.~\ref{alg:cap}) with the lowest PPL values. In doing so, we provide a mechanism for shifting the model's focus towards learning data blocks that were determined to be difficult.
    \end{enumerate}
    \item \textbf{Review}: Next, we reintroduce all of the removed data blocks and train the model by randomly sampling from the entire corpus for $p_3$ steps (line 13 in Alg.~\ref{alg:cap}). This ensures that we allow the model to review and revisit data blocks which it may have forgotten.  
\end{enumerate}

\begin{algorithm}
\caption{LFR Training Methodology}\label{alg:cap}
\begin{algorithmic}[1]
\Require Training dataset $D$, model $M$ with initial parameters $\theta_0$, hyperparameters $p_1$, $s_1$, $p_2$, $p_3$, $reps$, and $epochs$. 
\Ensure Minimization of Equation \ref{eq:nll}.

\State $PPLs, \theta_{p_1} \gets \textbf{Learn}(\theta_0, D, p_1)$
\For{$r=1,2,\ldots,reps$}
    \If{$epochs == 1$}
        \State $D_k \gets Cluster(D)$
        \State $Sort(PPLs, D_k)$
        \State $S_{sub} \gets (1-s_1)\times D_k$
        \State $S_1 \gets sample(S_{sub}, PPLs)$
        % \State $S_1  weighted_sampling$
        % \State $Sort(PPLs, D_k), S_1 \gets (100-s_1)\%D_k$
    \Else
        \State $Sort(PPLs, D)$
        \State $S_1 \gets (1-s_1)\times D$
    \EndIf
    \State $\theta_{p_2} \gets \textbf{Focus}(\theta_{p_1}, S_1, p_2)$
    \State $PPLs, \theta_{p_3} \gets \textbf{Review}(\theta_{p_2}, D, p_3)$
\EndFor
\newline $Return \ \theta$
\end{algorithmic}
\end{algorithm}

 Our training strategy is simple, intuitive and human-like. It gives the model an opportunity to learn from all of the data, prioritize and relearn forgotten data points, and review data samples from harder regions of the dataset more frequently than they would have been using random sampling.
 While the static clustering-based techniques~\cite{tirumala2023d4, abbas2023semdedup, kaddour2023minipile} presented in Section~\ref{prior_work} allow for accelerated training, they are not designed to suit the multi-descent training dynamics observed in Section~\ref{method} and require pretrained model embeddings to calculate distance metrics.
 % By dropping data points with high similarity (within the same cluster), the model is not given a chance to repeatedly learn harder data points and observe the multi-descent behaviour in Figure~\ref{fig:medium-forgetting}.
 Furthermore, prior methods including perplexity-based pruning methods~\cite{marion2023more} are static. Sections~\ref{ablation-dropped-retained} and \ref{app_ablation_extension} characterize the data blocks found easy and hard by the LLM, and demonstrate why static, clustering-based data selection methods achieve poor downstream task performance.
 Lastly, our approach does not require any pretrained models to obtain embeddings.
 Our focused training strategy allows the model to absorb harder information (data blocks with higher perplexity) faster, by presenting them more number of times. 

\section{Evaluation}\label{eval}
In this section, we present a comprehensive evaluation of LFR. We pretrain the Llama-2 models of sizes 120M, 300M, and 500M, and the GPT-2 models~\cite{Radford2019LanguageMA} of various sizes between 124M and 1.5B parameters. We use the SlimPajama~\cite{cerebras2023slimpajama} and OpenWebText dataset~\cite{openwebtext} and train from scratch using 4 NVIDIA A6000 GPUs (Llama-2), 4 AMD MI210 GPUs (GPT-2 345M-1.5B), 4 AMD MI100 GPUs (GPT-2 124M). Our pretraining experiments utilize a fully distributed data parallel (DDP) approach. All model configurations and training hyperparameters of our experiments are detailed in Section~\ref{app_training_details}.

We also evaluate our pretrained models on various downstream tasks from different domains in Sections~\ref{llama_and_gpt_performance} and \ref{comparison-prior}. Section~\ref{pruning-ablation} demonstrates the impact of the Focus and Review stages. Sections~\ref{llama-data} presents examples of data points from the SlimPajama dataset which were identified as easy and challenging by LFR. Sections~\ref{app_ablation_extension} and ~\ref{ablation-dropped-retained} presents an analysis on the data blocks dropped and retained by LFR in different phases of training across the different GPT-2 models. Together Sections~\ref{llama-data}, ~\ref{app_ablation_extension}, and ~\ref{ablation-dropped-retained} demonstrate that LLMs learn instructions and code after facts and anecdotes as illustrated in Figure~\ref{llm-training-timeline}. Finally, we conduct a sensitivity study in Section~\ref{sensitivity}, where we test LFR with different hyperparameters.

\begin{table*}[h]
\centering
    \begin{tabularx}{\textwidth}{|c|X|X|X|X|X|X|X|X|}
    \hline 
    Model  & Tokens & Arc\_C & Arc\_E & Boolq & HS & OBQA & Piqa & WG \\
            \hline
         120M-RS & 50B &  23.29& 39.06&53.15&34.30&30.50&62.58&51.54\\
         Pythia-160M & 300B & 20.1 & \textbf{44.0} &  40.0 & \textbf{35.82}& 29.59& 61.80 & 49.7 \\
         \textbf{120M-LFR} & 9.8B &  \textbf{23.61}&39.52&\textbf{54.86}&35.44&\textbf{30.56}&\textbf{63.21}&\textbf{53.88}\\
         \hline
         300M-RS &50B & 25.7 & 41.9 & 54.0 & 36.4 & 33.3 & 64.8 & 52.4 \\
         Pythia-410M & 300B & 30.3 & \textbf{47.1} & \textbf{55.3} & 40.1 & 36.2 & \textbf{67.2} & 53.4 \\
         \textbf{300M-LFR} & 9.8B &\textbf{36.61}&44.52&54.86&\textbf{48.44}&\textbf{40.56}&66.21&\textbf{56.88}\\
         \hline
         500M-RS &50B& 27.5 & 43.7 & 53.7 & 36.5 & 32.6 & 65.1 & 52.2\\
         Pythia-1.0B & 300B & 37.05 & \textbf{56.99} & \textbf{60.83} & 47.16 & 39.40 & \textbf{70.70} & 58.70 \\
         \textbf{500M-LFR} & 9.8B & \textbf{38.81} & 53.10 & 58.72 & \textbf{50.65} & \textbf{43.10} & 68.66 & \textbf{58.72}\\
         \hline         
    \end{tabularx}
    \caption{Zero-shot performance (acc\_norm for all except Winogrande and Boolq which use acc) on Common Sense Reasoning Tasks evaluated using LLM Evaluation Harness~\cite{eval-harness}. RS refers to the random sampling baseline, HS refers to HellaSwag, and WG refers to Winogrande. The model with the highest performance (measured by acc\_norm) is highlighted in bold. Notably, LFR models are trained using only 3.2\% of the tokens required to train Pythia models of comparable size, yet they achieve higher accuracy in 55.55\% of cases. Additionally, LFR models consistently outperform the random sampling baseline by a large margin, despite being trained on 19.6\% of the pretraining tokens.}
    \label{tab:llama-results}
\end{table*}
\subsection{LFR Configuration}
We pretrain the Llama models for a total of 100k steps and train on 9.8B tokens based on the Chinchilla scaling law~\cite{hoffmann2022trainingcomputeoptimallargelanguage}. All training hyperparameters are described in Section~\ref{app_training_details}. First, we Learn for 20k steps ($p_1 = 20k$). Next, we cluster the data and discard 57.2\% of the clusters, retaining only the 3 most challenging clusters out of 7 based on their $PPL$ values ($s_1 = 50$). We then apply the Focus stage for 60k steps ($p_2 = 60k$), prioritizing the retained high-PPL clusters. The clustering results using ($k=7$) are shown in Figure~\ref{fig:clustering}. It takes <10min to cluster which can be hidden by the high training latency. We provide a detailed analysis on the hierarchical clustering and the data points found easy and difficult in Section~\ref{eval}. Lastly, we Review the entire dataset for the last 20k steps ($p_3 = 20k$). 
In the case of the GPT-2 models, we Learn for 1 epoch ($p_1 = 1$), Focus on 50\% of the data for 1 epoch ($s_1 = 50, p_2 = 1$), Review the entire dataset for another epoch ($p_3 = 1$), and Focus on 30\% of the data for 5 epochs ($reps = 2, s_2 = 70, p_4 = 1$). 

We chose this configuration based on our limited pretraining budget and profiling in Section \ref{obs}, which showed that 78.5\% and 25\% of data samples are forgotten at least once during training for Llama and GPT-2 models, respectively. Figure \ref{fig:medium-forgetting} reveals that forgotten samples are typically forgotten multiple times, requiring an average of 4 presentations to be learned. For GPT-2, we use the first three phases to identify these samples and allocate 5 epochs focusing on 30\% of them in the final phase to ensure long-term retention. Additionally, all samples are reviewed through Phases 1 and 3, and Section \ref{sensitivity} evaluates LFR's sensitivity to hyperparameters $p_1$, $s_1$, $p_2$, $p_3$, and $reps$.

\subsection{Baselines}\label{baselines}
We evaluate the models pretrained using LFR with a comprehensive set of prior works and industry-standard checkpoints. They include:
\begin{enumerate}
    \item Industry-standard models: We compare the Llama-2 models trained through LFR with Pythia models~\cite{10.5555/3618408.3618510} of similar scale obtained from EleutherAI's Huggingface\footnote{https://huggingface.co/models?other=pythia}. These models have been trained on 300B tokens while the LFR models were trained on 9.8B tokens (3.2\% of the tokens). We compare the GPT-2 models pretrained through LFR for 40k iterations with the same GPT-2 architectures pretrained by OpenAI~\footnote{https://huggingface.co/openai-community} for 800k iterations (5\% of the training tokens). We use the same batch size as these industry-standard models (Section~\ref{app_training_details}) by adjusting the gradient accumulation steps and the per-device batch size. 
    \item Random Sampling: while the previous baselines ensures that we compare with industry-standard models, we also develop and compare LFR against the same models pretrained using random sampling with 5.10$\times$ and 20$\times$ more tokens than LFR for the Llama-2 and GPT-2 models respectively. This baseline enables LFR to produce higher quality models than those obtained through traditional autoregressive modeling when using much fewer tokens and training iterations.
    \item Prior works: We compare our training methodology with the dataset generated through DSIR~\cite{xie2023data} and MiniPile~\cite{kaddour2023minipile}.
\end{enumerate}

\subsection{Performance on Downstream Tasks}\label{llama_and_gpt_performance}
We perform single-shot evaluations on common-sense reasoning tasks using Llama-2 models trained through LFR, comparing the accuracy norm with the baselines in Table~\ref{tab:llama-results}. Note that the LFR models are trained on only 3.2\% of the tokens used to train the Pythia models and 19.6\% of the tokens used for the random-sampling baseline. Additionally, the Pythia models have between $1.3\times$ and $2.0\times$ more parameters. Despite this, LFR models outperform the Pythia models on 55.55\% of the datasets and consistently surpass the random-sampling baseline.

We test the GPT-2 models on language modeling tasks and compare with the OpenAI baseline in Table~\ref{tab:langmodel-results} by measuring the PPL. Note that our models are trained on 5\% of the training tokens as compared with the OpenAI models, further validating that data quality is more important than quantity. We find that the PPL reduction obtained by LFR increase as the dataset size increases (from WikiText-2 to 1BW). Also, smaller models show a larger $PPL$ reduction by using LFR than larger models. On average, using our approach, perplexity was reduced by 4.92, 3.26, 2.17, and 1.40 for the GPT-2 124M, 345M, 774M, and 1.5B models, respectively. 

\begin{table*}[h]
\centering
    \begin{tabularx}{0.97\textwidth}{|c|X|X|X|X|}
    \hline 
    Model  & WikiText-2 & WikiText-103 & LAMBADA & 1BW \\
            \hline
         124M-OpenAI (800k iters) &  22.1 & 31.58 & 18 & 39.18 \\
         124M-RS (40k iters)  & 23.32 & 23.42 & 17.71 & 39.49\\ 
         124M-LFR (40k iters) &  19.81 & 22.49 & 16.61 & 32.27 \\
         \hline
         345M-OpenAI (800k iters) & 19.82 & 22.05 & 14.26 & 29.95 \\
         345M-RS (40k iters)  & 21.11 & 21.8 & 14.84 & 30.66\\
         345M-LFR (40k iters) &  16.31 & 17.48 & 13.7 & 25.52 \\
         \hline
         774M-OpenAI (800k iters) & 15.93 & 18.53 & 13.74 & 26.52 \\
         774M-RS (40k iters)  & 16.71 & 18.89 & 14.10 & 28.56 \\
         774M-LFR (40k iters) &  15.11 & 14.58 & 12.51 & 23.83 \\
         \hline
         1.5B-OpenAI (800k iters) & 13.80 & 16.59 & 12.15 &  23.87\\
         1.5B-LFR (40k iters) &  13.10 &  14.37 & 11.23 &  22.09 \\
         \hline         
    \end{tabularx}
    \caption{$PPL$ results for language modeling datasets across model sizes. Here, $N$-OpenAI refers to the OpenAI baseline (trained for 800k iterations), $N$-RS refers to the random sampling baseline (trained for 40k iterations), and $N$-LFR refers to our proposed training pedagogy (trained for 40k iterations), where $N$ is the number of model parameters.  }
    \label{tab:langmodel-results}
\end{table*}

We also test the LFR-trained models on standard benchmarks from the translation, question-answering, world knowledge, and problem solving domains in Table~\ref{tab:model_performance_gpt2}. LFR models trained with 20$\times$ fewer training iterations achieves better performance than models trained using random sampling. Details of each of the datasets is provided in Section~\ref{app_training_details}.

\begin{table*}[h]
\centering
\begin{tabularx}{\textwidth}{|l|l|X|X|X|X|X|X|X|}
\hline
\multirow{2}{*}{Model} & \multirow{2}{*}{Iters} & WMT & NQ & \multicolumn{5}{c|}{MMLU}\\
\cline{5-9}
& & (BLEU)& (Acc) & STEM (Acc) & HM (Acc) & SS (Acc) & Other (Acc) & Avg. (Acc)\\
\hline
1.5B OpenAI & 800k & 11.5 & 4.1 & 24.5 & 24.8 & \textbf{24.0} & \textbf{27.8} & 25.3 \\
\textbf{1.5B LFR} & 40k& \textbf{11.8} & \textbf{4.61} & \textbf{26.1} & \textbf{27.2} & 23.8 & 25.1 & \textbf{25.5} \\
\hline
\end{tabularx}
\caption{LFR-trained GPT-2 models evaluated on translation (WMT-14~\cite{wmt-14}), question-answering (Natural Questions~\cite{kwiatkowski-etal-2019-natural}), and world knowledge and problem solving (MMLU~\cite{hendryckstest2021} domains using the BLEU scores and accuracy metrics. Note that NQ refers to Natural Questions, HM refers to Humanities, SS refers to Social Sciences, Other refers to business, health, and other miscellaneous topics, and Avg. refers to the average accuracy across all 57 subjects in MMLU. We compare our 1.5B parameter model with those trained by OpenAI for 20$\times$ more training iterations. The model with the superior performance is highlighted in bold. }
\label{tab:model_performance_gpt2}
\end{table*}

\subsection{Comparison with Prior Work}\label{comparison-prior}
\begin{table}[h]
    \centering
    \adjustbox{width=\columnwidth}{
    \begin{tabular}{|c|c|c|c|c|}
    \hline
        & WikiText-2 & WikiText-103 & Lambada & 1BW \\
        \hline
         DSIR & 22.6 & 25.29 & 17.34 & 35.33 \\
         MiniPile & 150 & 322.1 & 73.61 & 459.2 \\
         LFR & 19.81 & 22.49 & 16.61 & 32.27 \\
         \hline
    \end{tabular}}
    \caption{$PPL$ achieved by LFR and prior work on data selection for accelerated LLM training.}
    \label{tab:comparison-prior-work}
\end{table}
In this section, we use the 124M parameter model to compare LFR with DSIR~\cite{xie2023data} and MiniPile~\cite{kaddour2023minipile}, which propose importance resampling and cluster-based data sampling respectively. In the case of DSIR, we use the OpenWebText dataset as the raw dataset and set the target dataset as WikiText-103 for selecting a subset of 1M documents ($\sim$6GB). It took a total of 2.7 hours for the data selection process on 8 AMD EPYC 7V13 CPU cores operating at 2.45GHz with 64GB RAM. For the MiniPile baseline, we pretrain the GPT-2 124M model for 40k iterations with the dataset available on Huggingface~\citep{minipile-huggingface}. 

Table~\ref{tab:comparison-prior-work} displays the downstream task perplexity. LFR significantly outperforms prior work on data selection and allows the pretrained LLM to generalize better to new finetuning objectives, by focusing and revising different parts of the training corpus at regular intervals.
\begin{figure}[h]
  \includegraphics[width=\columnwidth]{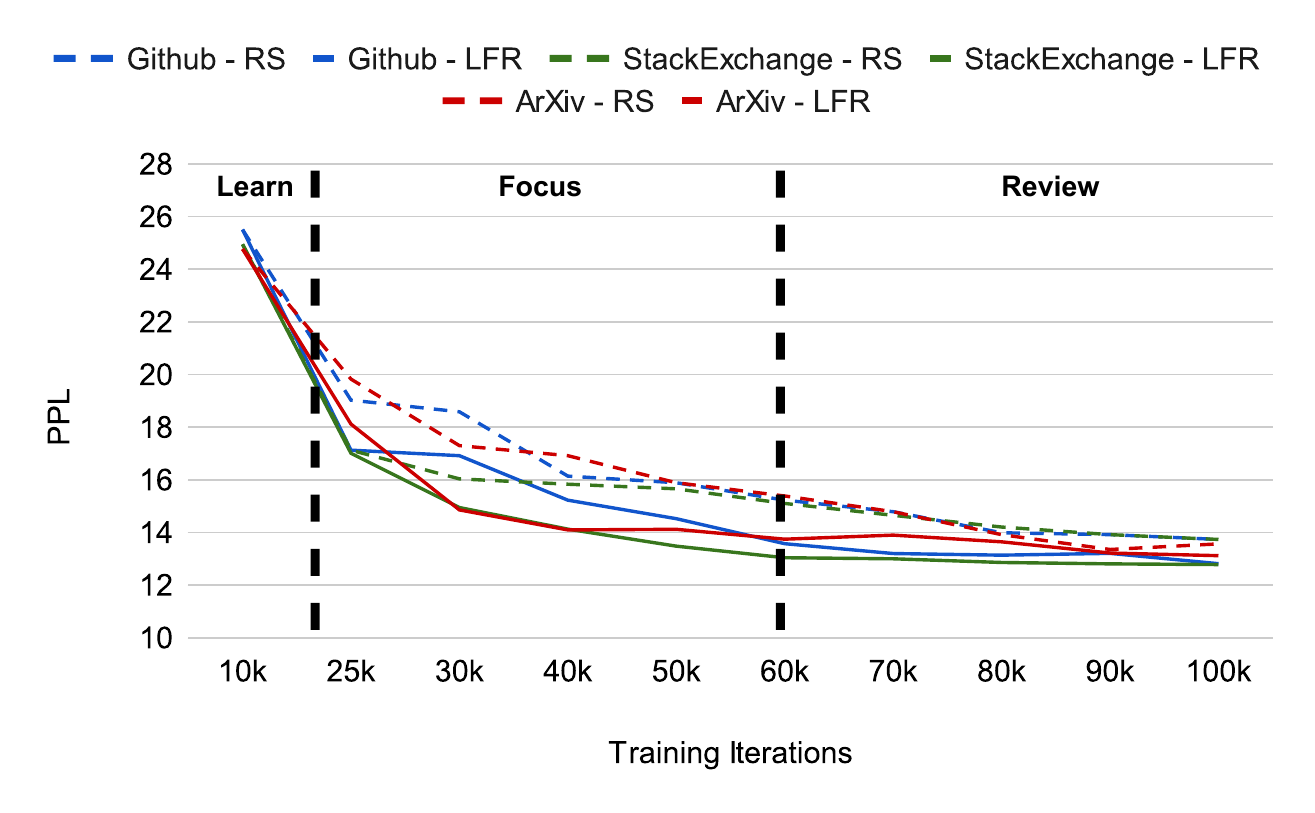}
  \caption{PPL values are tracked at different training iterations for the clusters identified as challenging and prioritized during the Focus stage of LFR. The dotted line represents the PPL values for the same clusters when trained with random sampling (RS). Notably, LFR facilitates accelerated learning of these challenging data points between 20k and 60k iterations (the Focus stage), whereas random sampling consistently results in higher PPL values throughout.}
  \label{fig:focus-clusters}
\end{figure}

\begin{figure}[h!]
  \includegraphics[width=\columnwidth]{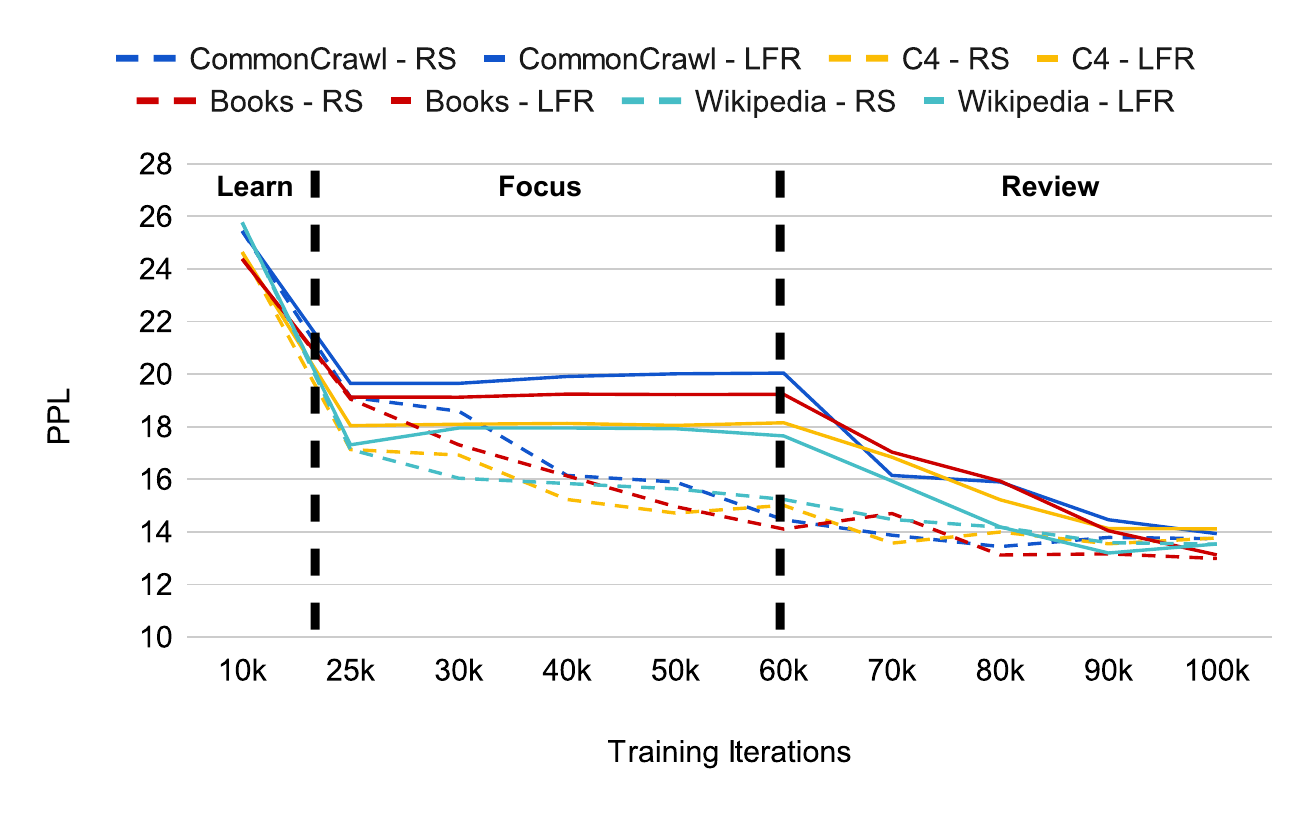}
  \caption{PPL values are tracked at different training iterations for the clusters identified as easy, discarded during the Focus stage, and reintroduced during the Review phase. The dotted line represents the PPL values for the same clusters when trained with random sampling (RS). Notably, we demonstrate that models forget the data points discarded during training, unless reintroduced to the training corpus as in the case of LFR.}
  \label{fig:review-clusters}
\end{figure}

\subsection{Ablation Study}\label{pruning-ablation}
In this section, our goal is to understand the impacts of the Focus and Review stages of LFR and exploring more aggressive data selection strategies by varying the hyperparameters $p_1, s_1, p_2, p_3,$ and $reps$.

\subsubsection{Impact of Focus}
Consider training the Llama-2 300M parameter model on the SlimPajama dataset, which comprises of seven sub-datasets sourced from CommonCrawl, Github, C4, Books, Wikipedia, StackExchange, and ArXiv. During the Focus stage, LFR employs weighted sampling from the three most challenging clusters while discarding clusters with the lowest perplexity (PPL). Additionally, within the retained clusters, LFR performs hierarchical sampling by prioritizing regions with higher PPL, further refining the data selection process. LFR classifies the Github, StackExchange, and ArXiv clusters as more challenging at 20k iterations, than the other four data sources. Figure~\ref{fig:focus-clusters} demonstrates the training dynamics of these challenging data points. Note that LFR (solid line) enables the accelerated learning of challenging parts of the dataset as opposed to random-sampling based training (dotted line). 

Section~\ref{llama-data} provides examples of data considered challenging and easy from each of the sources in the SlimPajama dataset.

\subsubsection{Impact of Review}
Next, we study the impact of the Review phase on the data points identified as simple, and discarded during the Focus stage. Note that these samples are reintroduced to the training process in LFR, unlike prior work in data selection. Figure~\ref{fig:review-clusters} demonstrates the importance of the Review phase by plotting the PPL values for the easy data points for LFR (solid line) and random sampling (dotted line). The model forgets data points which are discarded during training (solid line is higher than the dotted line in the Focus stage). This is because the model is focusing on more challenging clusters during this phase. Therefore, the Review phase is essential to achieving performant models, and provides LFR a distinct edge over other data selection methodologies evaluated in Section~\ref{comparison-prior}. Please refer to Section~\ref{llama-data} for examples of data samples identified as easy and difficult by LFR.

\section{Conclusion}
We present LFR, a new training paradigm which proposes a mechanism for models to learn, focus, revisit and review complex data during training. Our method is inspired by the effective learning methods used by humans. LFR achieves consistently lower perplexity and higher accuracy on downstream task performance over traditional autoregressive modeling-based methods while using 20x fewer training iterations. We also demonstrate that LLMs first learn anecdotal and conversational data, before being able to retain factual information. 

\section{Acknowledgements}
This work is partially supported by the PRISM Center under the JUMP 2.0 Program, AMD HACC Program, and NSF 2048280, 2331966, 2325121, 2244760. The authors would like to thank AMD for their generous support provided through the AMD AI \& HPC Fund Research Award.
\section{Limitations and Ethical Considerations}
LFR presents the following directions for future work:
\begin{enumerate}
    \item LFR is evaluated on models up to 1.5B parameters using the OpenWebText dataset, constrained by our compute resources. With the clear success on models of such scale, we hope to inspire researchers to validate such focused learning approaches for different model families, and domains. 
    \item The sensitivity study in Section~\ref{pruning-ablation} reveals that the hyperparameters selected in Section~\ref{method} have a large impact on the performance of the trained model. Due to our limited compute budget, we are unable to present more comprehensive hyperparameter tuning experiments than those presented in Section~\ref{pruning-ablation}.
    
\end{enumerate}

\bibliography{custom}

\begin{thebibliography}{45}
\providecommand{\natexlab}[1]{#1}

\bibitem[{arc()}]{arc_challenge}

\newblock {Arc Challenge Dataset}.
\newblock \url{https://huggingface.co/datasets/allenai/ai2_arc}.

\bibitem[{arc()}]{arc_easy}
 b.
\newblock {Arc Easy Dataset}.
\newblock \url{https://huggingface.co/datasets/allenai/ai2_arc}.

\bibitem[{boo()}]{bookcorpus}

\newblock {BookCorpus Dataset}.
\newblock \url{https://huggingface.co/datasets/bookcorpus/bookcorpus}.

\bibitem[{boo()}]{boolq}
 b.
\newblock {BoolQ Dataset}.
\newblock \url{https://huggingface.co/datasets/google/boolq}.

\bibitem[{gpt()}]{gpt4-openai}

\newblock {GPT-4 Cost Estimation}.
\newblock \url{https://en.wikipedia.org/wiki/GPT-4#:~:text=Sam%20Altman%20stated%20that%20the,was%20more%20than%20%24100%20million.}

\bibitem[{hel()}]{hellaswag}

\newblock {HellaSwag Dataset}.
\newblock \url{https://huggingface.co/datasets/DatologyAI/hellaswag}.

\bibitem[{min()}]{minipile-huggingface}

\newblock {MiniPile Dataset}.
\newblock \url{https://huggingface.co/datasets/JeanKaddour/minipile}.

\bibitem[{Ope()}]{OpenBookQA}

\newblock {OpenBookQA Dataset}.
\newblock \url{https://huggingface.co/datasets/allenai/openbookqa}.

\bibitem[{ope()}]{openwebtext}

\newblock {OpenWebText Dataset}.
\newblock \url{https://huggingface.co/datasets/Skylion007/openwebtext}.

\bibitem[{Piq()}]{Piqa}

\newblock {PIQA Dataset}.
\newblock \url{https://huggingface.co/datasets/ybisk/piqa}.

\bibitem[{wik()}]{wikitext}

\newblock {WikiText Dataset}.
\newblock \url{https://huggingface.co/datasets/Salesforce/}.

\bibitem[{win()}]{winogrande}

\newblock {WinoGrande Dataset}.
\newblock \url{https://huggingface.co/datasets/allenai/winogrande}.

\bibitem[{wmt()}]{wmt-14}

\newblock {WMT-14 Hugging Face Dataset}.
\newblock \url{https://huggingface.co/datasets/wmt/wmt14}.

\bibitem[{lla(2024)}]{llama3}
 2024.
\newblock {Meta Llama 3}.
\newblock \url{https://ai.meta.com/blog/meta-llama-3/}.

\bibitem[{Abbas et~al.(2023)Abbas, Tirumala, Simig, Ganguli, and Morcos}]{abbas2023semdedup}
Amro Abbas, Kushal Tirumala, Dániel Simig, Surya Ganguli, and Ari~S. Morcos. 2023.
\newblock \href {https://arxiv.org/abs/2303.09540} {Semdedup: Data-efficient learning at web-scale through semantic deduplication}.
\newblock \emph{Preprint}, arXiv:2303.09540.

\bibitem[{Artetxe et~al.(2018)Artetxe, Labaka, Agirre, and Cho}]{artetxe2018unsupervised}
Mikel Artetxe, Gorka Labaka, Eneko Agirre, and Kyunghyun Cho. 2018.
\newblock \href {https://openreview.net/forum?id=Sy2ogebAW} {Unsupervised neural machine translation}.
\newblock In \emph{International Conference on Learning Representations}.

\bibitem[{Biderman et~al.(2023)Biderman, Prashanth, Sutawika, Schoelkopf, Anthony, Purohit, and Raff}]{biderman2023emergent}
Stella Biderman, USVSN~Sai Prashanth, Lintang Sutawika, Hailey Schoelkopf, Quentin Anthony, Shivanshu Purohit, and Edward Raff. 2023.
\newblock \href {https://arxiv.org/abs/2304.11158} {Emergent and predictable memorization in large language models}.
\newblock \emph{Preprint}, arXiv:2304.11158.

\bibitem[{Biderman et~al.(2023)Biderman, Schoelkopf, Anthony, Bradley, O'Brien, Hallahan, Khan, Purohit, Prashanth, Raff, Skowron, Sutawika, and Van Der~Wal}]{10.5555/3618408.3618510}
Stella Biderman, Hailey Schoelkopf, Quentin Anthony, Herbie Bradley, Kyle O'Brien, Eric Hallahan, Mohammad~Aflah Khan, Shivanshu Purohit, USVSN~Sai Prashanth, Edward Raff, Aviya Skowron, Lintang Sutawika, and Oskar Van Der~Wal. 2023b.
\newblock Pythia: a suite for analyzing large language models across training and scaling.
\newblock In \emph{Proceedings of the 40th International Conference on Machine Learning}, ICML'23. JMLR.org.

\bibitem[{Brown et~al.(2020)Brown, Mann, Ryder, Subbiah, Kaplan, Dhariwal, Neelakantan, Shyam, Sastry, Askell, Agarwal, Herbert-Voss, Krueger, Henighan, Child, Ramesh, Ziegler, Wu, Winter, Hesse, Chen, Sigler, Litwin, Gray, Chess, Clark, Berner, McCandlish, Radford, Sutskever, and Amodei}]{NEURIPS2020_1457c0d6}
Tom Brown, Benjamin Mann, Nick Ryder, Melanie Subbiah, Jared~D Kaplan, Prafulla Dhariwal, Arvind Neelakantan, Pranav Shyam, Girish Sastry, Amanda Askell, Sandhini Agarwal, Ariel Herbert-Voss, Gretchen Krueger, Tom Henighan, Rewon Child, Aditya Ramesh, Daniel Ziegler, Jeffrey Wu, Clemens Winter, Chris Hesse, Mark Chen, Eric Sigler, Mateusz Litwin, Scott Gray, Benjamin Chess, Jack Clark, Christopher Berner, Sam McCandlish, Alec Radford, Ilya Sutskever, and Dario Amodei. 2020.
\newblock \href {https://proceedings.neurips.cc/paper_files/paper/2020/file/1457c0d6bfcb4967418bfb8ac142f64a-Paper.pdf} {Language models are few-shot learners}.
\newblock In \emph{Advances in Neural Information Processing Systems}, volume~33, pages 1877--1901. Curran Associates, Inc.

\bibitem[{Chelba et~al.(2014)Chelba, Mikolov, Schuster, Ge, Brants, Koehn, and Robinson}]{chelba2014billion}
Ciprian Chelba, Tomas Mikolov, Mike Schuster, Qi~Ge, Thorsten Brants, Phillipp Koehn, and Tony Robinson. 2014.
\newblock \href {https://arxiv.org/abs/1312.3005} {One billion word benchmark for measuring progress in statistical language modeling}.
\newblock \emph{Preprint}, arXiv:1312.3005.

\bibitem[{Gao et~al.(2020)Gao, Biderman, Black, Golding, Hoppe, Foster, Phang, He, Thite, Nabeshima, Presser, and Leahy}]{gao2020pile}
Leo Gao, Stella Biderman, Sid Black, Laurence Golding, Travis Hoppe, Charles Foster, Jason Phang, Horace He, Anish Thite, Noa Nabeshima, Shawn Presser, and Connor Leahy. 2020.
\newblock \href {https://arxiv.org/abs/2101.00027} {{The Pile: An 800GB Dataset of Diverse Text for Language Modeling}}.
\newblock \emph{Preprint}, arXiv:2101.00027.

\bibitem[{Gao et~al.(2024)Gao, Tow, Abbasi, Biderman, Black, DiPofi, Foster, Golding, Hsu, Le~Noac'h, Li, McDonell, Muennighoff, Ociepa, Phang, Reynolds, Schoelkopf, Skowron, Sutawika, Tang, Thite, Wang, Wang, and Zou}]{eval-harness}
Leo Gao, Jonathan Tow, Baber Abbasi, Stella Biderman, Sid Black, Anthony DiPofi, Charles Foster, Laurence Golding, Jeffrey Hsu, Alain Le~Noac'h, Haonan Li, Kyle McDonell, Niklas Muennighoff, Chris Ociepa, Jason Phang, Laria Reynolds, Hailey Schoelkopf, Aviya Skowron, Lintang Sutawika, Eric Tang, Anish Thite, Ben Wang, Kevin Wang, and Andy Zou. 2024.
\newblock \href {https://doi.org/10.5281/zenodo.12608602} {A framework for few-shot language model evaluation}.

\bibitem[{Hendrycks et~al.(2021)Hendrycks, Burns, Basart, Zou, Mazeika, Song, and Steinhardt}]{hendryckstest2021}
Dan Hendrycks, Collin Burns, Steven Basart, Andy Zou, Mantas Mazeika, Dawn Song, and Jacob Steinhardt. 2021.
\newblock Measuring massive multitask language understanding.
\newblock \emph{Proceedings of the International Conference on Learning Representations (ICLR)}.

\bibitem[{Hoffmann et~al.(2022)Hoffmann, Borgeaud, Mensch, Buchatskaya, Cai, Rutherford, de~Las~Casas, Hendricks, Welbl, Clark, Hennigan, Noland, Millican, van~den Driessche, Damoc, Guy, Osindero, Simonyan, Elsen, Rae, Vinyals, and Sifre}]{hoffmann2022trainingcomputeoptimallargelanguage}
Jordan Hoffmann, Sebastian Borgeaud, Arthur Mensch, Elena Buchatskaya, Trevor Cai, Eliza Rutherford, Diego de~Las~Casas, Lisa~Anne Hendricks, Johannes Welbl, Aidan Clark, Tom Hennigan, Eric Noland, Katie Millican, George van~den Driessche, Bogdan Damoc, Aurelia Guy, Simon Osindero, Karen Simonyan, Erich Elsen, Jack~W. Rae, Oriol Vinyals, and Laurent Sifre. 2022.
\newblock \href {https://arxiv.org/abs/2203.15556} {Training compute-optimal large language models}.
\newblock \emph{Preprint}, arXiv:2203.15556.

\bibitem[{Jiang et~al.(2023)Jiang, Sablayrolles, Mensch, Bamford, Chaplot, de~las Casas, Bressand, Lengyel, Lample, Saulnier, Lavaud, Lachaux, Stock, Scao, Lavril, Wang, Lacroix, and Sayed}]{jiang2023mistral}
Albert~Q. Jiang, Alexandre Sablayrolles, Arthur Mensch, Chris Bamford, Devendra~Singh Chaplot, Diego de~las Casas, Florian Bressand, Gianna Lengyel, Guillaume Lample, Lucile Saulnier, Lélio~Renard Lavaud, Marie-Anne Lachaux, Pierre Stock, Teven~Le Scao, Thibaut Lavril, Thomas Wang, Timothée Lacroix, and William~El Sayed. 2023.
\newblock \href {https://arxiv.org/abs/2310.06825} {Mistral 7b}.
\newblock \emph{Preprint}, arXiv:2310.06825.

\bibitem[{Kaddour(2023)}]{kaddour2023minipile}
Jean Kaddour. 2023.
\newblock \href {https://arxiv.org/abs/2304.08442} {{The MiniPile Challenge for Data-Efficient Language Models}}.
\newblock \emph{Preprint}, arXiv:2304.08442.

\bibitem[{Kwiatkowski et~al.(2019)Kwiatkowski, Palomaki, Redfield, Collins, Parikh, Alberti, Epstein, Polosukhin, Devlin, Lee, Toutanova, Jones, Kelcey, Chang, Dai, Uszkoreit, Le, and Petrov}]{kwiatkowski-etal-2019-natural}
Tom Kwiatkowski, Jennimaria Palomaki, Olivia Redfield, Michael Collins, Ankur Parikh, Chris Alberti, Danielle Epstein, Illia Polosukhin, Jacob Devlin, Kenton Lee, Kristina Toutanova, Llion Jones, Matthew Kelcey, Ming-Wei Chang, Andrew~M. Dai, Jakob Uszkoreit, Quoc Le, and Slav Petrov. 2019.
\newblock \href {https://doi.org/10.1162/tacl_a_00276} {Natural questions: A benchmark for question answering research}.
\newblock \emph{Transactions of the Association for Computational Linguistics}, 7:452--466.

\bibitem[{Lin et~al.(2024)Lin, Gou, Gong, Liu, yelong shen, Xu, Lin, Yang, Jiao, Duan, and Chen}]{lin2024not}
Zhenghao Lin, Zhibin Gou, Yeyun Gong, Xiao Liu, yelong shen, Ruochen Xu, Chen Lin, Yujiu Yang, Jian Jiao, Nan Duan, and Weizhu Chen. 2024.
\newblock \href {https://openreview.net/forum?id=0NMzBwqaAJ} {Not all tokens are what you need for pretraining}.
\newblock In \emph{The Thirty-eighth Annual Conference on Neural Information Processing Systems}.

\bibitem[{Marion et~al.(2023)Marion, Üstün, Pozzobon, Wang, Fadaee, and Hooker}]{marion2023more}
Max Marion, Ahmet Üstün, Luiza Pozzobon, Alex Wang, Marzieh Fadaee, and Sara Hooker. 2023.
\newblock \href {https://arxiv.org/abs/2309.04564} {{When Less is More: Investigating Data Pruning for Pretraining LLMs at Scale}}.
\newblock \emph{Preprint}, arXiv:2309.04564.

\bibitem[{OpenAI et~al.(2024)OpenAI, Achiam, Adler, Agarwal, Ahmad, Akkaya, Aleman, Almeida, Altenschmidt, Altman, Anadkat, Avila, Babuschkin, Balaji, Balcom, Baltescu, Bao, Bavarian, Belgum, Bello, Berdine, Bernadett-Shapiro, Berner, Bogdonoff, Boiko, Boyd, Brakman, Brockman, Brooks, Brundage, Button, Cai, Campbell, Cann, Carey, Carlson, Carmichael, Chan, Chang, Chantzis, Chen, Chen, Chen, Chen, Chen, Chess, Cho, Chu, Chung, Cummings, Currier, Dai, Decareaux, Degry, Deutsch, Deville, Dhar, Dohan, Dowling, Dunning, Ecoffet, Eleti, Eloundou, Farhi, Fedus, Felix, Fishman, Forte, Fulford, Gao, Georges, Gibson, Goel, Gogineni, Goh, Gontijo-Lopes, Gordon, Grafstein, Gray, Greene, Gross, Gu, Guo, Hallacy, Han, Harris, He, Heaton, Heidecke, Hesse, Hickey, Hickey, Hoeschele, Houghton, Hsu, Hu, Hu, Huizinga, Jain, Jain, Jang, Jiang, Jiang, Jin, Jin, Jomoto, Jonn, Jun, Kaftan, Łukasz Kaiser, Kamali, Kanitscheider, Keskar, Khan, Kilpatrick, Kim, Kim, Kim, Kirchner, Kiros, Knight, Kokotajlo, Łukasz Kondraciuk,
  Kondrich, Konstantinidis, Kosic, Krueger, Kuo, Lampe, Lan, Lee, Leike, Leung, Levy, Li, Lim, Lin, Lin, Litwin, Lopez, Lowe, Lue, Makanju, Malfacini, Manning, Markov, Markovski, Martin, Mayer, Mayne, McGrew, McKinney, McLeavey, McMillan, McNeil, Medina, Mehta, Menick, Metz, Mishchenko, Mishkin, Monaco, Morikawa, Mossing, Mu, Murati, Murk, Mély, Nair, Nakano, Nayak, Neelakantan, Ngo, Noh, Ouyang, O'Keefe, Pachocki, Paino, Palermo, Pantuliano, Parascandolo, Parish, Parparita, Passos, Pavlov, Peng, Perelman, de~Avila Belbute~Peres, Petrov, de~Oliveira~Pinto, Michael, Pokorny, Pokrass, Pong, Powell, Power, Power, Proehl, Puri, Radford, Rae, Ramesh, Raymond, Real, Rimbach, Ross, Rotsted, Roussez, Ryder, Saltarelli, Sanders, Santurkar, Sastry, Schmidt, Schnurr, Schulman, Selsam, Sheppard, Sherbakov, Shieh, Shoker, Shyam, Sidor, Sigler, Simens, Sitkin, Slama, Sohl, Sokolowsky, Song, Staudacher, Such, Summers, Sutskever, Tang, Tezak, Thompson, Tillet, Tootoonchian, Tseng, Tuggle, Turley, Tworek, Uribe, Vallone,
  Vijayvergiya, Voss, Wainwright, Wang, Wang, Wang, Ward, Wei, Weinmann, Welihinda, Welinder, Weng, Weng, Wiethoff, Willner, Winter, Wolrich, Wong, Workman, Wu, Wu, Wu, Xiao, Xu, Yoo, Yu, Yuan, Zaremba, Zellers, Zhang, Zhang, Zhao, Zheng, Zhuang, Zhuk, and Zoph}]{openai2024gpt4}
OpenAI, Josh Achiam, Steven Adler, Sandhini Agarwal, Lama Ahmad, Ilge Akkaya, Florencia~Leoni Aleman, Diogo Almeida, Janko Altenschmidt, Sam Altman, Shyamal Anadkat, Red Avila, Igor Babuschkin, Suchir Balaji, Valerie Balcom, Paul Baltescu, Haiming Bao, Mohammad Bavarian, Jeff Belgum, Irwan Bello, Jake Berdine, Gabriel Bernadett-Shapiro, Christopher Berner, Lenny Bogdonoff, Oleg Boiko, Madelaine Boyd, Anna-Luisa Brakman, Greg Brockman, Tim Brooks, Miles Brundage, Kevin Button, Trevor Cai, Rosie Campbell, Andrew Cann, Brittany Carey, Chelsea Carlson, Rory Carmichael, Brooke Chan, Che Chang, Fotis Chantzis, Derek Chen, Sully Chen, Ruby Chen, Jason Chen, Mark Chen, Ben Chess, Chester Cho, Casey Chu, Hyung~Won Chung, Dave Cummings, Jeremiah Currier, Yunxing Dai, Cory Decareaux, Thomas Degry, Noah Deutsch, Damien Deville, Arka Dhar, David Dohan, Steve Dowling, Sheila Dunning, Adrien Ecoffet, Atty Eleti, Tyna Eloundou, David Farhi, Liam Fedus, Niko Felix, Simón~Posada Fishman, Juston Forte, Isabella Fulford, Leo
  Gao, Elie Georges, Christian Gibson, Vik Goel, Tarun Gogineni, Gabriel Goh, Rapha Gontijo-Lopes, Jonathan Gordon, Morgan Grafstein, Scott Gray, Ryan Greene, Joshua Gross, Shixiang~Shane Gu, Yufei Guo, Chris Hallacy, Jesse Han, Jeff Harris, Yuchen He, Mike Heaton, Johannes Heidecke, Chris Hesse, Alan Hickey, Wade Hickey, Peter Hoeschele, Brandon Houghton, Kenny Hsu, Shengli Hu, Xin Hu, Joost Huizinga, Shantanu Jain, Shawn Jain, Joanne Jang, Angela Jiang, Roger Jiang, Haozhun Jin, Denny Jin, Shino Jomoto, Billie Jonn, Heewoo Jun, Tomer Kaftan, Łukasz Kaiser, Ali Kamali, Ingmar Kanitscheider, Nitish~Shirish Keskar, Tabarak Khan, Logan Kilpatrick, Jong~Wook Kim, Christina Kim, Yongjik Kim, Jan~Hendrik Kirchner, Jamie Kiros, Matt Knight, Daniel Kokotajlo, Łukasz Kondraciuk, Andrew Kondrich, Aris Konstantinidis, Kyle Kosic, Gretchen Krueger, Vishal Kuo, Michael Lampe, Ikai Lan, Teddy Lee, Jan Leike, Jade Leung, Daniel Levy, Chak~Ming Li, Rachel Lim, Molly Lin, Stephanie Lin, Mateusz Litwin, Theresa Lopez, Ryan
  Lowe, Patricia Lue, Anna Makanju, Kim Malfacini, Sam Manning, Todor Markov, Yaniv Markovski, Bianca Martin, Katie Mayer, Andrew Mayne, Bob McGrew, Scott~Mayer McKinney, Christine McLeavey, Paul McMillan, Jake McNeil, David Medina, Aalok Mehta, Jacob Menick, Luke Metz, Andrey Mishchenko, Pamela Mishkin, Vinnie Monaco, Evan Morikawa, Daniel Mossing, Tong Mu, Mira Murati, Oleg Murk, David Mély, Ashvin Nair, Reiichiro Nakano, Rajeev Nayak, Arvind Neelakantan, Richard Ngo, Hyeonwoo Noh, Long Ouyang, Cullen O'Keefe, Jakub Pachocki, Alex Paino, Joe Palermo, Ashley Pantuliano, Giambattista Parascandolo, Joel Parish, Emy Parparita, Alex Passos, Mikhail Pavlov, Andrew Peng, Adam Perelman, Filipe de~Avila Belbute~Peres, Michael Petrov, Henrique~Ponde de~Oliveira~Pinto, Michael, Pokorny, Michelle Pokrass, Vitchyr~H. Pong, Tolly Powell, Alethea Power, Boris Power, Elizabeth Proehl, Raul Puri, Alec Radford, Jack Rae, Aditya Ramesh, Cameron Raymond, Francis Real, Kendra Rimbach, Carl Ross, Bob Rotsted, Henri Roussez,
  Nick Ryder, Mario Saltarelli, Ted Sanders, Shibani Santurkar, Girish Sastry, Heather Schmidt, David Schnurr, John Schulman, Daniel Selsam, Kyla Sheppard, Toki Sherbakov, Jessica Shieh, Sarah Shoker, Pranav Shyam, Szymon Sidor, Eric Sigler, Maddie Simens, Jordan Sitkin, Katarina Slama, Ian Sohl, Benjamin Sokolowsky, Yang Song, Natalie Staudacher, Felipe~Petroski Such, Natalie Summers, Ilya Sutskever, Jie Tang, Nikolas Tezak, Madeleine~B. Thompson, Phil Tillet, Amin Tootoonchian, Elizabeth Tseng, Preston Tuggle, Nick Turley, Jerry Tworek, Juan Felipe~Cerón Uribe, Andrea Vallone, Arun Vijayvergiya, Chelsea Voss, Carroll Wainwright, Justin~Jay Wang, Alvin Wang, Ben Wang, Jonathan Ward, Jason Wei, CJ~Weinmann, Akila Welihinda, Peter Welinder, Jiayi Weng, Lilian Weng, Matt Wiethoff, Dave Willner, Clemens Winter, Samuel Wolrich, Hannah Wong, Lauren Workman, Sherwin Wu, Jeff Wu, Michael Wu, Kai Xiao, Tao Xu, Sarah Yoo, Kevin Yu, Qiming Yuan, Wojciech Zaremba, Rowan Zellers, Chong Zhang, Marvin Zhang, Shengjia
  Zhao, Tianhao Zheng, Juntang Zhuang, William Zhuk, and Barret Zoph. 2024.
\newblock \href {https://arxiv.org/abs/2303.08774} {{GPT-4 Technical Report}}.
\newblock \emph{Preprint}, arXiv:2303.08774.

\bibitem[{Paperno et~al.(2016)Paperno, Kruszewski, Lazaridou, Pham, Bernardi, Pezzelle, Baroni, Boleda, and Fern{\'a}ndez}]{paperno-etal-2016-lambada}
Denis Paperno, Germ{\'a}n Kruszewski, Angeliki Lazaridou, Ngoc~Quan Pham, Raffaella Bernardi, Sandro Pezzelle, Marco Baroni, Gemma Boleda, and Raquel Fern{\'a}ndez. 2016.
\newblock \href {https://doi.org/10.18653/v1/P16-1144} {The {LAMBADA} dataset: Word prediction requiring a broad discourse context}.
\newblock In \emph{Proceedings of the 54th Annual Meeting of the Association for Computational Linguistics (Volume 1: Long Papers)}, pages 1525--1534, Berlin, Germany. Association for Computational Linguistics.

\bibitem[{Paul et~al.(2021)Paul, Ganguli, and Dziugaite}]{Paul2021DeepLO}
Mansheej Paul, Surya Ganguli, and Gintare~Karolina Dziugaite. 2021.
\newblock \href {https://api.semanticscholar.org/CorpusID:235898952} {Deep learning on a data diet: Finding important examples early in training}.
\newblock In \emph{Neural Information Processing Systems}.

\bibitem[{Penedo et~al.(2023)Penedo, Malartic, Hesslow, Cojocaru, Cappelli, Alobeidli, Pannier, Almazrouei, and Launay}]{crawl}
Guilherme Penedo, Quentin Malartic, Daniel Hesslow, Ruxandra Cojocaru, Alessandro Cappelli, Hamza Alobeidli, Baptiste Pannier, Ebtesam Almazrouei, and Julien Launay. 2023.
\newblock \href {https://arxiv.org/abs/2306.01116} {{The RefinedWeb Dataset for Falcon LLM: Outperforming Curated Corpora with Web Data, and Web Data Only}}.
\newblock \emph{Preprint}, arXiv:2306.01116.

\bibitem[{Radford et~al.(2019)Radford, Wu, Child, Luan, Amodei, and Sutskever}]{Radford2019LanguageMA}
Alec Radford, Jeff Wu, Rewon Child, David Luan, Dario Amodei, and Ilya Sutskever. 2019.
\newblock \href {https://api.semanticscholar.org/CorpusID:160025533} {Language models are unsupervised multitask learners}.

\bibitem[{Raffel et~al.(2023)Raffel, Shazeer, Roberts, Lee, Narang, Matena, Zhou, Li, and Liu}]{raffel2023exploring}
Colin Raffel, Noam Shazeer, Adam Roberts, Katherine Lee, Sharan Narang, Michael Matena, Yanqi Zhou, Wei Li, and Peter~J. Liu. 2023.
\newblock \href {https://arxiv.org/abs/1910.10683} {Exploring the limits of transfer learning with a unified text-to-text transformer}.
\newblock \emph{Preprint}, arXiv:1910.10683.

\bibitem[{Smolen et~al.(2016)Smolen, Zhang, and Byrne}]{Smolen_2016}
Paul Smolen, Yili Zhang, and John~H. Byrne. 2016.
\newblock \href {https://doi.org/10.1038/nrn.2015.18} {The right time to learn: mechanisms and optimization of spaced learning}.
\newblock \emph{Nature Reviews Neuroscience}, 17(2):77–88.

\bibitem[{Soboleva et~al.(2023)Soboleva, Al-Khateeb, Myers, Steeves, Hestness, and Dey}]{cerebras2023slimpajama}
Daria Soboleva, Faisal Al-Khateeb, Robert Myers, Jacob~R Steeves, Joel Hestness, and Nolan Dey. 2023.
\newblock \href {https://huggingface.co/datasets/cerebras/SlimPajama-627B} {{SlimPajama: A 627B token cleaned and deduplicated version of RedPajama}}.
\newblock \url{https://cerebras.ai/blog/slimpajama-a-627b-token-cleaned-and-deduplicated-version-of-redpajama}.

\bibitem[{Tabibian et~al.(2019)Tabibian, Upadhyay, De, Zarezade, Schölkopf, and Gomez-Rodriguez}]{pnas-spacerep}
Behzad Tabibian, Utkarsh Upadhyay, Abir De, Ali Zarezade, Bernhard Schölkopf, and Manuel Gomez-Rodriguez. 2019.
\newblock \href {https://doi.org/10.1073/pnas.1815156116} {Enhancing human learning via spaced repetition optimization}.
\newblock \emph{Proceedings of the National Academy of Sciences}, 116(10):3988--3993.

\bibitem[{Team(2024)}]{geminiteam2024geminifamilyhighlycapable}
Gemini Team. 2024.
\newblock \href {https://arxiv.org/abs/2312.11805} {Gemini: A family of highly capable multimodal models}.
\newblock \emph{Preprint}, arXiv:2312.11805.

\bibitem[{Tirumala et~al.(2023)Tirumala, Simig, Aghajanyan, and Morcos}]{tirumala2023d4}
Kushal Tirumala, Daniel Simig, Armen Aghajanyan, and Ari~S. Morcos. 2023.
\newblock \href {https://arxiv.org/abs/2308.12284} {{D4: Improving LLM Pretraining via Document De-Duplication and Diversification}}.
\newblock \emph{Preprint}, arXiv:2308.12284.

\bibitem[{Touvron et~al.(2023)Touvron, Lavril, Izacard, Martinet, Lachaux, Lacroix, Rozière, Goyal, Hambro, Azhar, Rodriguez, Joulin, Grave, and Lample}]{touvron2023llama}
Hugo Touvron, Thibaut Lavril, Gautier Izacard, Xavier Martinet, Marie-Anne Lachaux, Timothée Lacroix, Baptiste Rozière, Naman Goyal, Eric Hambro, Faisal Azhar, Aurelien Rodriguez, Armand Joulin, Edouard Grave, and Guillaume Lample. 2023.
\newblock \href {https://arxiv.org/abs/2302.13971} {Llama: Open and efficient foundation language models}.
\newblock \emph{Preprint}, arXiv:2302.13971.

\bibitem[{Wang et~al.(2024)Wang, Yang, Huang, Jiao, Yang, Jiang, Majumder, and Wei}]{wang2024text}
Liang Wang, Nan Yang, Xiaolong Huang, Binxing Jiao, Linjun Yang, Daxin Jiang, Rangan Majumder, and Furu Wei. 2024.
\newblock \href {https://arxiv.org/abs/2212.03533} {Text embeddings by weakly-supervised contrastive pre-training}.
\newblock \emph{Preprint}, arXiv:2212.03533.

\bibitem[{Xia et~al.(2022)Xia, Artetxe, Zhou, Lin, Pasunuru, Chen, Zettlemoyer, and Stoyanov}]{Xia2022TrainingTO}
Mengzhou Xia, Mikel Artetxe, Chunting Zhou, Xi~Victoria Lin, Ramakanth Pasunuru, Danqi Chen, Luke Zettlemoyer, and Ves Stoyanov. 2022.
\newblock \href {https://api.semanticscholar.org/CorpusID:254877112} {Training trajectories of language models across scales}.
\newblock \emph{ArXiv}, abs/2212.09803.

\bibitem[{Xie et~al.(2023)Xie, Santurkar, Ma, and Liang}]{xie2023data}
Sang~Michael Xie, Shibani Santurkar, Tengyu Ma, and Percy Liang. 2023.
\newblock \href {https://arxiv.org/abs/2302.03169} {Data selection for language models via importance resampling}.
\newblock \emph{Preprint}, arXiv:2302.03169.

\bibitem[{Zhang et~al.(2022)Zhang, Roller, Goyal, Artetxe, Chen, Chen, Dewan, Diab, Li, Lin, Mihaylov, Ott, Shleifer, Shuster, Simig, Koura, Sridhar, Wang, and Zettlemoyer}]{Zhang2022OPTOP}
Susan Zhang, Stephen Roller, Naman Goyal, Mikel Artetxe, Moya Chen, Shuohui Chen, Christopher Dewan, Mona~T. Diab, Xian Li, Xi~Victoria Lin, Todor Mihaylov, Myle Ott, Sam Shleifer, Kurt Shuster, Daniel Simig, Punit~Singh Koura, Anjali Sridhar, Tianlu Wang, and Luke Zettlemoyer. 2022.
\newblock \href {https://api.semanticscholar.org/CorpusID:248496292} {Opt: Open pre-trained transformer language models}.
\newblock \emph{ArXiv}, abs/2205.01068.

\end{thebibliography}
% \pagebreak
% \clearpage
\appendix

% \newpage

% \input{appendix_latex}
\section{Appendix}
\label{sec:appendix}
\subsection{Experiment Details}\label{app_training_details}
\textbf{Datasets}
The datasets used for our experiments in Section \ref{eval} are detailed below:
\begin{enumerate}
    \item ARC-Challenge~\cite{arc_challenge}: A subset of the AI2 Reasoning Challenge with 2,590 challenging multiple-choice science questions designed to test advanced reasoning and knowledge.
    \item ARC-Easy~\cite{arc_easy}: A subset of the AI2 Reasoning Challenge with 5,117 relatively easier multiple-choice science questions focusing on basic reasoning and recall.
    \item BoolQ~\cite{boolq}: A dataset of 16,000+ boolean (yes/no) questions paired with passages, requiring models to infer answers from supporting evidence. 
    \item HellaSwag~\cite{hellaswag}: A dataset with 70,000+ multiple-choice questions focused on commonsense reasoning and contextual understanding, particularly in describing scenarios.
    \item OpenBookQA~\cite{OpenBookQA}: A multiple-choice question-answering dataset with 5,957 questions requiring knowledge retrieval from a science "open book" and commonsense reasoning.
    \item PIQA~\cite{Piqa}: A physical commonsense reasoning dataset with 20,000+ binary-choice questions about everyday situations and physical interactions.
    \item Winogrande~\cite{winogrande}: A dataset with 44,000+ sentence pairs designed to test commonsense reasoning through pronoun disambiguation challenges.
    \item WikiText~\cite{wikitext}: the WikiText language modeling dataset consists of 100M tokens extracted from Wikipedia articles with high rating. It features two different variants, namely, WikiText-2 and WikiText-103 which differ in the number of tokens and vocabulary size. WikiText-2 consists of 2M tokens and a vocabulary size of 33k whereas WikiText-103 is larger with 103M tokens and a vocabulary size of 267k. 
    \item LAMBADA~\cite{paperno-etal-2016-lambada}: the LAMBADA dataset is extracted from the BookCorpus dataset~\cite{bookcorpus} and contains 10k passages. This dataset is useful for testing the ability of an LLM to capture long-range dependencies in text. The objective of this model is to predict the final word in a set of sentences, where humans need at least 50 tokens of context to accurately anticipate the word. 
    \item One Billion Word Benchmark~\cite{chelba2014billion} (1BW): this dataset contains one billion words extracted from the WMT 2011 News Crawl data and is used to measure progress in statistical language modeling. 
    \item WMT-14 French-English Translation \cite{artetxe2018unsupervised}: This dataset contains 36 million training sentence pairs for english to french translation. The test set, which is used for evaluation purposes, consists of 3,003 sentence pairs. 
    \item Natural Questions \cite{kwiatkowski-etal-2019-natural}: This dataset contains question-answer pairs from Google Search and Wikipedia-based annotations. The training, validation, and test sets consist of 307,372, 7,830, and 7,842 examples. 
\end{enumerate}

% \citep{Radford2019LanguageMA} demonstrates that these datasets are out-of-domain for the WebText dataset using Bloom filters containing 8-grams of the training tokens. While we have not conducted a similar study for the OpenWebText dataset, since both datasets are scraped and pre-processed using a similar method, we estimate that the percentages of 8 grams overlap with the training set will be similar. 

\textbf{Models}: 
Tables \ref{tab:llama_model_configs} and \ref{tab:llama_pretraining_hyperparams} describes the different model configurations and pretraining hyperparameters used in LFR for the Llama-2 models.

\begin{table}[h]
    \centering
    \begin{tabular}{|c|c|c|c|}
    \hline
        & 120M & 300M & 500M \\  
        \hline
        Layers & 12 & 12 & 11  \\
        \#Heads & 12 & 16 & 32 \\
        n\_embd & 768 & 1024 & 2048 \\
    \hline
    \end{tabular}
    \caption{Number of layers, attention heads, and the embedding dimensions in the Llama-2 models used for pretraining. }
    \label{tab:llama_model_configs}
\end{table}

\begin{table}[h]
    \centering
    \begin{tabularx}{\columnwidth}{|X|X|}
    \hline
        Parameter & Value \\
        \hline
        Context Length & 1024  \\
        Embedding Dimension & (768, 1024, 2048) \\
        Total Iterations & 100,000 \\
        Effective Batch Size & 768 \\
        Block Size & 4096 \\
        Weight Decay & 1.00E-1 \\
        Adam $\beta_1$ & 0.90 \\
        Adam $\beta_2$ & 0.95 \\
        Warmup Iterations & 8000 \\
        Minimum Learning Rate & 4.00E-5 \\
        Maximum Learning Rate & 4.00E-04 \\
        Learning Rate Schedule & Cosine Decay \\
        Learning Rate Decay Iterations & 100,000 \\
        GPUs & (4x NVIDIA A6000) \\
        \hline
    \end{tabularx}
    \caption{Pretraining hyperparameters for the Llama-2 120M-500M parameter models. Parameters with multiple values (e.g. Embedding dimensions, batch size, gradient accumulation steps, and GPUs) specified in brackets are for the 120M, 300M, and 500M parameter models respectively.}
    \label{tab:llama_pretraining_hyperparams}
\end{table}

Tables \ref{tab:model-configs} and \ref{tab:pretraining_hyperparams} describes the different model configurations and pretraining hyperparameters used in LFR for the GPT-2 models. 

\begin{table}[h]
    \centering
    \begin{tabular}{|c|c|c|c|c|}
    \hline
        & 124M & 355M & 774M & 1.5B \\  
        \hline
        Layers & 12 & 24 & 36 & 48  \\
        \#Heads & 12 & 16 & 20 & 25 \\
        n\_embd & 768 & 1024 & 1280 & 1600 \\
    \hline
    \end{tabular}
    \caption{Number of layers, attention heads, and the embedding dimensions in the GPT-2 models used for pretraining. }
    \label{tab:model-configs}
\end{table}
\textbf{Pretraining}: Table \ref{tab:pretraining_hyperparams} shows the hyperparameters for pretraining the GPT-2 124M-1.5B parameter models. 

\begin{table}[h]
    \centering
    \begin{tabularx}{\columnwidth}{|X|X|}
    \hline
        Parameter & Value \\
        \hline
        Context Length & 1024  \\
        Embedding Dimension & (768, 1024, 1280, 1600) \\
        Total Iterations & 40000 \\
        Effective Batch Size & 512 \\
        Batch Size & (16, 16, 8, 4) \\
        Gradient Accumulation Steps & (32, 32, 64, 128) \\
        Block Size & 1024 \\
        Weight Decay & 1.00E-01 \\
        Adam $\beta_1$ & 0.9 \\
        Adam $\beta_2$ & 0.95 \\
        Warmup Iterations & 2000 \\
        Minimum Learning Rate & 6.00E-05 \\
        Maximum Learning Rate & 6.00E-04 \\
        Learning Rate Schedule & Linear \\
        Learning Rate Decay Iterations & 40000 \\
        GPUs & (4xMI100, 4xMI210, 4xMI210, 4xMI210) \\
        \hline
    \end{tabularx}
    \caption{Pretraining hyperparameters for the GPT-2 124M-1.5B parameter models. Parameters with multiple values (e.g. Embedding dimensions, batch size, gradient accumulation steps, and GPUs) specified in brackets are for the 124M, 345M, 774M, and 1.5B parameter models respectively.}
    \label{tab:pretraining_hyperparams}
\end{table}
Note that OpenAI pretrained the GPT-2 models using a batch size of 512. Due to insufficient GPU memory, we adjust the number of gradient accumulation steps to achieve the same effective batch size of 512.

% \textbf{Training Runtime}
% Table~\ref{tab:runtime_in_days} describes the runtime in days per pretraining experiment (40k iterations) for the different model configurations. 

% \begin{table}[h]
%     \centering
%     \begin{tabular}{|c|c|c|}
%     \hline
%          Model & \#Days & GPUs  \\
%          \hline
%          GPT-2 124M & 2.2 & 4 AMD MI100 \\
%          GPT-2 345M & 8.8 & 4 AMD MI210 \\
%          GPT-2 812M & 14.1 & 4 AMD MI210 \\
%          GPT-2 1.5B & 20 & 4 AMD MI210 \\
%          \hline
%     \end{tabular}
%     \caption{Pretraining experiments runtime and GPU configurations used.}
%     \label{tab:runtime_in_days}
% \end{table}

% \begin{table*}[h]
%     \centering
%     \begin{tabularx}{\textwidth}{|c|X|X|X|X|X|X|X|}
%     \hline 
%     Model & Arc\_C & Arc\_E & Boolq & HS & OBQA & Piqa & WG \\
%     \hline
%     Llama 300M (HF) (50B tokens) & 23.3 & 39.3 & 53.2
%  & 34.3 & 30.6 & 64.6 & 51.5 \\
%  \hline
%   Bert-large uncased (336M) & 25.7 & 25.1 & 40.9 & 24.5 & 26.2 & 47.7 & 49.8\\
%   \hline

%     Llama 300M (LFR) (9B tokens) & 23.6 & 39.5 &  54.9 & 35.4 & 30.6 & 63.2 & 53.9\\
    
% \hline
%     \end{tabularx}
%     \caption{Zero-shot evaluation on problem solving, commonsense reasoning, and question-answering tasks.}
% \end{table*}
\textbf{Finetuning}: We use all the same hyperparameters as pretraining, except for the following:
\begin{enumerate}
    \item Learning rate: 3.00E-5 
    \item Learning rate schedule: Constant
    \item Total iterations: 50
\end{enumerate}

\subsection{Llama Pretraining - Data Importance}\label{llama-data}

\begin{figure}
    \centering
    \includegraphics[width=\columnwidth]{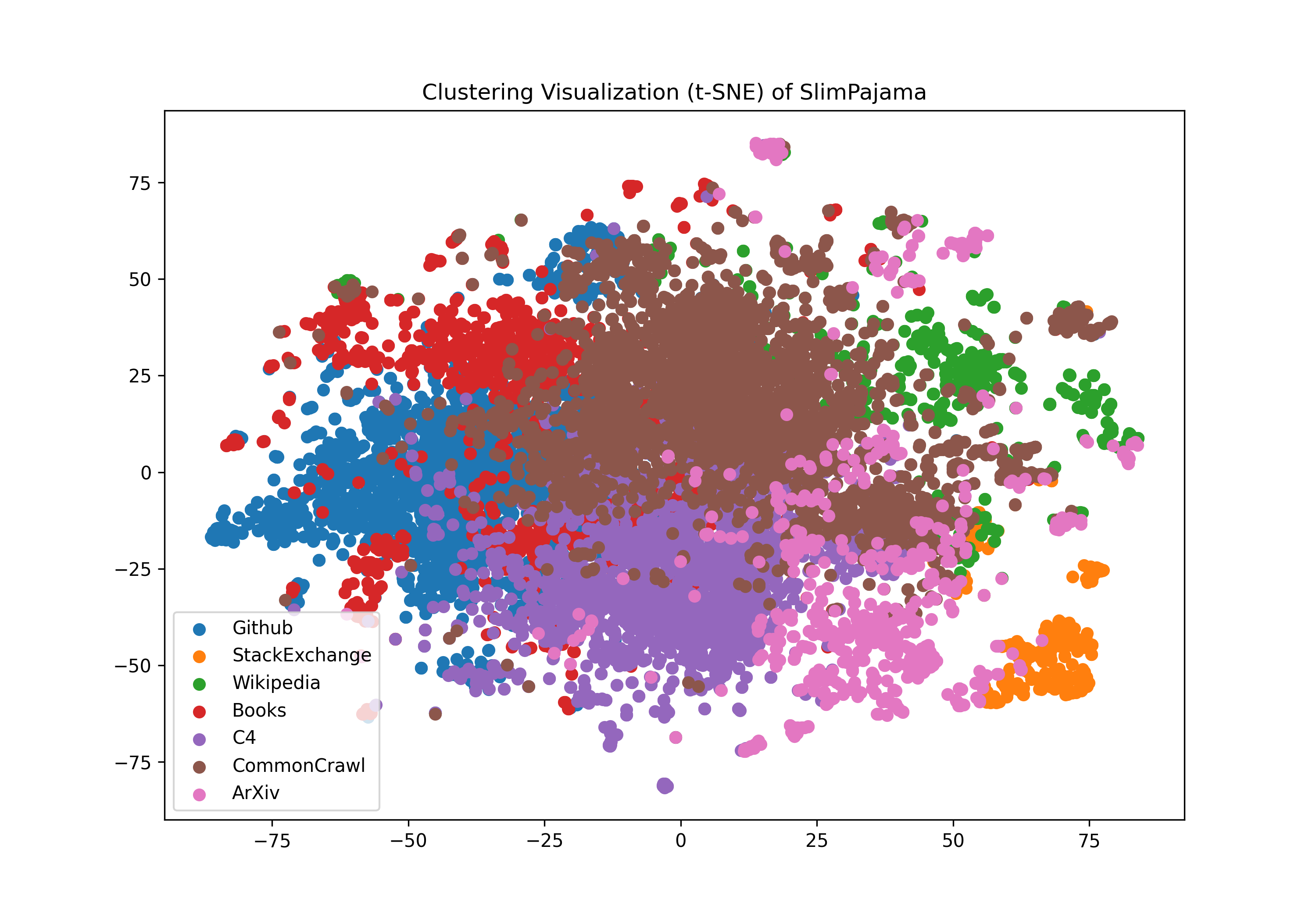}
    \caption{Clustering the data embeddings from the SlimPajama dataset using the Llama-300M model at the 50k training step.}
    \label{fig:clustering}
\end{figure}

In this section, we study the data points identified as easy and challenging by LFR when pretraining with the SlimPajama dataset. Listing~\ref{lst:trcdbgtrace} provides an example of a code snippet from Github classified as easy by LFR, and discarded in the Focus stage of the Llama model training. Listing~\ref{lst:difficult} provides an example of a data sample retained from the Github cluster. Note that this code is more complex, presents an opportunity to the model to improve its coding capabilities as opposed to the variable declarations in Listing~\ref{lst:trcdbgtrace}.

\begin{lstlisting}[language=java, breaklines=true, caption={Code snippet classified as easy by LFR, primarily consisting of variable declarations. As seen from the code, it contributes minimally to enhancing the model's coding capabilities.}]
package frclibj;

import edu.wpi.first.wpilibj.Timer;

public class TrcDbgTrace
{
    public static final String ESC_PREFIX       = "\u001b[";
    public static final String ESC_SUFFIX       = "m";
    public static final String ESC_SEP          = ";";

    public static final String SGR_RESET        = "0";
    public static final String SGR_BRIGHT       = "1";
    public static final String SGR_DIM          = "2";
    public static final String SGR_ITALIC       = "3";
    public static final String SGR_UNDERLINE    = "4";
    public static final String SGR_BLINKSLOW    = "5";
    public static final String SGR_BLINKFAST    = "6";
    public static final String SGR_REVERSE      = "7";
    public static final String SGR_HIDDEN       = "8";
    public static final String SGR_CROSSEDOUT   = "9";

    public static final String ESC_NORMAL       = ESC_PREFIX;
}
\end{lstlisting}
% \captionof{listing}{Code snippet classified as easy by LFR, primarily consisting of variable declarations. As seen from the code, it contributes minimally to enhancing the model's coding capabilities.}
\label{lst:trcdbgtrace}

\begin{lstlisting}[breaklines, caption={Code snippet classified as challenging by LFR. This code consists of a function which executes an Oracle query and returns a scalar value. As seen from the code, it contributes significantly to enhancing the model's coding capabilities as compared with Listing~\ref{lst:trcdbgtrace}.}]{csharp}
/// <summary>
/// Executes an Oracle query that returns a single scalar value as the result.
/// </summary>
/// <param name="commandText">The Oracle query to execute</param>
/// <param name="parameters">Optional parameters to pass to the query</param>
/// <returns>The result of the query as an object</returns>
public object QueryValue(string commandText, IEnumerable parameters)
{
    object result;

    if (String.IsNullOrEmpty(commandText))
    {
        throw new ArgumentException("Command text cannot be null or empty.");
    }

    try
    {
        ensureConnectionOpen();
        var command = createCommand(commandText, parameters);
        result = command.ExecuteScalar();
    }
    finally
    {
        ensureConnectionClosed();
    }

    return result;
}

\end{lstlisting}
% \captionof{listing}{Code snippet classified as challenging by LFR. This code consists of a function which executes an Oracle query and returns a scalar value. As seen from the code, it contributes significantly to enhancing the model's coding capabilities as compared with Listing~\ref{lst:trcdbgtrace}.}
\label{lst:difficult}
% \vspace{1cm}

Similarly, we also provide examples of question-answer pairs from StackExchange which were discarded and retained in the Focus stage of the Llama pretraining in Listings~\ref{lst:easy-stackex} and \ref{lst:difficult-stackex} respectively.

\begin{lstlisting}[breaklines, caption={Question-answer pair from StackExchange classified as easy by LFR. The question revolves around a process in PayPal which does not contribute as much to the answering capability or world knowledge of the model.}]{text}
Q: PayPal IPN $_POST['txn_id'] not set. I'm using the PayPal sandbox to do a subscribe button, and then when I get the IPN response for a subscription or a subscription cancellation $_POST['txn_id'] is never set.
So I don't know how to identify transactions to only accept unique ones. 
Thanks!
EDIT: for example, all the info that I have in POST for a subscr_cancel are:
amount1, amount3, address_status, subscr_date, payer_id, address_street, mc_amount1, mc_amount3, charset, address_zip, first_name, reattempt, address_country_code, address_name, notify_version, subscr_id, custom, payer_status, business, address_country, address_city, verify_sign, payer_email, btn_id, last_name, address_state, receiver_email, recurring, txn_type, item_name, mc_currency, residence_country, test_ipn, period1, period3, correlation_id.

A: According to Table 2. Summary of subscription variables:
For subscription variables, the transaction ID (txn_id) is only available for USD Payment and Multi-Currency Payment transaction types (txn_type).

As expected, PayPal will not send the txn_id to your IPN for the transaction type, subscr_cancel, and will only send txn_id if the transaction type is subscr_payment.

For further explanation on which variables are sent to your IPN URL based on your transaction, please check out IPN and PDT Variables.

Have you checked $_REQUEST['txn_id'] as this may be sent to your server via GET.
\end{lstlisting}
% \captionof{listing}{Question-answer pair from StackExchange classified as easy by LFR. The question revolves around a process in PayPal which does not contribute as much to the answering capability or world knowledge of the model.}
\label{lst:easy-stackex}

\begin{lstlisting}[breaklines, caption={Question-answer pair from StackExchange classified as challenging by LFR. The question revolves around solving an ODE which contributes more to the learning of the model than Listing~\ref{lst:easy-stackex}. }]{text}
Q: Passing additional iteration-dependent inputs to ode45
I'm trying to solve a differential equation using the ode45 function. Consider the following code,
[t1,X2] = ode45(@(t,x)fun(t,x,C1,C2,C3,C4),t0,X01);

where parameters C1, C2, C3, and C4 are column vectors, which should be available to the function that ode45 is referring to (fun.m). 
I want the values to change after every iteration, so for example, at the beginning the entry of C1 I want is C1(1), in the next iteration it's C1(2), etc. 
How can I implement that?

A: You may have noticed that the official docs are not too helpful in this scenario (as they pretty much force you to use global variables - which is doable, but discouraged). 
Instead, I'll show you how this can be done with classes and function handles. Consider the following:

classdef SimpleQueue < handle
  %SIMPLEQUEUE A simple FIFO data structure.

  properties (Access = private)
    data
    position
  end

  methods (Access = public)
    function obj = SimpleQueue(inputData)
      %SIMPLEQUEUE Construct an instance of this class
      obj.data = inputData;
      rewind(obj);
    end % constructor

    function out = pop(obj, howMany)
      %POP return the next howMany elements.
      if nargin < 2 
        howMany = 1; % default amount of values to return
      end
      finalPosition = obj.position + howMany;
      if finalPosition > numel(obj.data)
        error('Too many elements requested!');
      end      
      out = obj.data(obj.position + 1 : obj.position + howMany);
      obj.position = finalPosition;      
    end % pop

    function [] = rewind(obj)
      %REWIND restarts the element tracking
      % Subsequent calls to pop() shall return elements from the beginning.
      obj.position = 0;
    end % rewind
  end % methods  
end % classdef

How to use this? Simple:
C1q = SimpleQueue(C1);
C2q = SimpleQueue(C2);
C3q = SimpleQueue(C3);
C4q = SimpleQueue(C4);

[t1,X2] = ode45(@(t,x)fun(t,x,@C1q.pop,@C2q.pop,
@C3q.pop,@C4q.pop),t0,X01);

As you can see, inside fun we use C1q() instead of C1.
\end{lstlisting}
% \captionof{listing}{Question-answer pair from StackExchange classified as challenging by LFR. The question revolves around solving an ODE which contributes more to the learning of the model than Listing~\ref{lst:easy-stackex}. }
\label{lst:difficult-stackex}

% \subsection{Sensitivity Study}\label{ablation}
\subsection{Sensitivity Study}\label{sensitivity}

% \subsection{Sensitivity Study} \label{pruning-ablation}
In this section, our goal is to understand the effects of more aggressive focus, revision, and learning strategies than the training strategy presented in Section~\ref{method}. Here, we vary the values of hyperparameters $p_1, s_1, p_2, p_3,$ and $reps$ and study the effects on the downstream task perplexity. Note that the GPT-2 models used a four phase training process.
Specifically, we aim to answer the following two questions using the GPT-2 models:

\begin{enumerate}
\item What is the impact of not reintroducing the discarded data samples?
\item What is the impact of the degree of pruning in Phases 2 and 4? 
\end{enumerate}

To answer the first question, we pretrain a 124M parameter GPT-2 model without the reintroduction of data blocks in Phase 3, and use the reduced subset from Phase 2 for the rest of the training. 
% with the following strategy: (1) Learn through random sampling for one epoch and record perplexities of data blocks (2) Sort and discard the 50\% of data blocks with the lowest perplexities for the previous training phase. Focus, and train on randomly sampled data batches from this reduced subset for the rest of the epochs (until 40k iterations). 
Then, we finetune for downstream language modeling tasks similarly and compared the perplexities using LFR in Table~\ref{tab:aggressive}. This training strategy which removes Phase 3, is labeled as \texttt{no-reintro}.
Next, to answer the second question, we pretrain a 124M parameter GPT-2 model using LFR but increase the degree of pruning in Phase 2 from 50\% to 70\%, i.e., reduce the training subset to 30\% of the original size.
This aggressive training strategy is labeled as \texttt{aggr-2}.

We observe that both aggressive training strategies do not work as well as the original method. However, we continue to explore more automated ways of deciding the training schedule for different model families as part of our future work.

\begin{table}[h]
    \centering
     \adjustbox{max width=\columnwidth}{
    \begin{tabular}{|c|c|c|c|c|}
    \hline
    Model & WikiText-2 & WikiText-103 & LAMBADA & 1BW \\
    \hline
    \texttt{no-reintro} & 23.24 & 25.76 & 17.27 & 36.02 \\
    \texttt{aggr-2} & 23.91 & 27.00 & 21.11 & 38.62 \\
    LFR & 19.81 & 22.49 & 16.61 & 32.27 \\
    \hline
    \end{tabular}}
    \caption{Downstream task perplexities with more aggressive training strategies.}
    \label{tab:aggressive}
\end{table}

\subsection{Analysis on Dropped and Retained Data Blocks - GPT-2}\label{ablation-dropped-retained}
% TODO FIGURES NOT APPEAR
\begin{figure*}[h]
  \includegraphics[width=\textwidth]{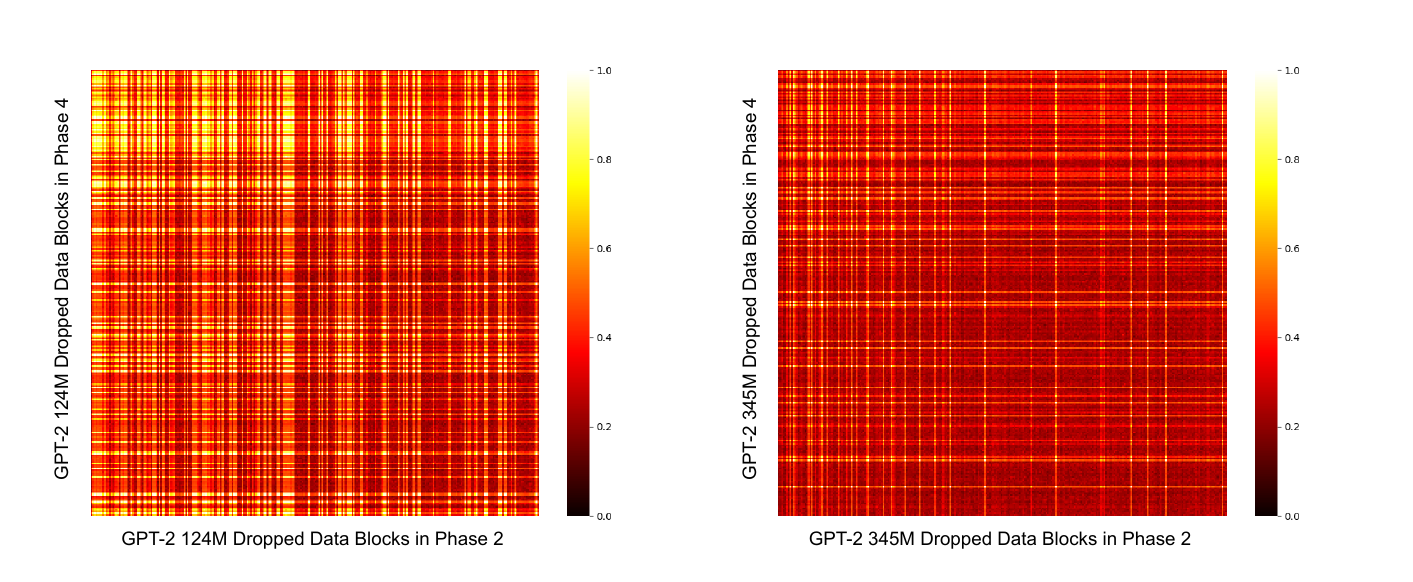}
  \caption{Cosine similarity heatmaps for dropped data blocks during phases 2 and 4 of pretraining for the GPT-2 124M (right) and 345M (left) models. The smaller model displays greater similarity in dropped data blocks over time (lighter color), indicating that it remained uncertain about similar data points than the larger model.}
  \label{fig:time_scale}
\end{figure*}
In this section, our goal is to characterize the data points retained and dropped during pretraining by LFR in Phases 2 and 4~(Sec.~\ref{method}) across the training time and model size. Specifically, we aim to answer the following questions:

\begin{enumerate}
\item What types of data blocks are learned earlier in the training process compared to those learned later?
\item Are similar data blocks considered learned and dropped in Phases 2 and 4?
% Are the dropped data blocks the same in both phases, i.e., does a data block's learnability change over time?
\item Are the dropped data blocks similar across model sizes?
% , i.e., are certain data blocks difficult to learn for a particular family of models?
\item Are the data blocks dropped similar to those retained at any given training phase?
% Are there similarities in the data blocks dropped and retained at training phases 2 and 4?
\end{enumerate}

To answer the first question, we printed out the texts dropped and retained at different training phases. Tables~\ref{tab:dropped_time} and \ref{tab:dropped_time_small} show text blocks dropped in Phases 2 and 4 by the 345M and 124M parameter models, while Tables~\ref{tab:dropped_time_retained} and \ref{tab:dropped_time_small_retained} show data blocks retained. By reading through the texts, we notice that the model first learned conversations and personal anecdotes, before being able to retain factual information. We provide a more detailed analysis of the learning process in Section~\ref{app_ablation_extension}.

% Tables~\ref{tab:dropped_time} and \ref{tab:dropped_time_small} provides examples of text blocks dropped in Phases 2 and 4 by the 345M and 124M parameter models respectively. Similarly, Tables~\ref{tab:dropped_time_retained} and \ref{tab:dropped_time_small_retained}
In order to answer questions 2-4, we recorded only the IDs of dropped data blocks during Phases 2 and 4 for both the GPT-2 124M and GPT-2 345M models, totaling 4 lists of dropped IDs. We then load the recorded data blocks and embed them into a higher dimensional space using the GPT-2 tokenizer.
Considering that there is a total of 8.7M data blocks (9B tokens divided into blocks of 1024 tokens), we cluster the embeddings using $k$-means clustering with $k=270$ to reduce the analysis space and complexity.
Finally, for each model, we compute the cosine similarity for all combinations of the embeddings of dropped data blocks across training phases and visualize them using a heatmap. These heatmaps plot the cosine similarity values (ranging between 0 and 1) such that lighter values (closer to 1) indicate higher similarity.
% We also calculate their mean, standard deviation, and variance.
%Figure~\ref{fig:time_scale} depicts a heatmap of the cosine similarity matrices across time scales for the GPT-2 124M and 345M models.
%The mean, standard deviation, and variance for the cosine similarity matrices in Figure \ref{fig:time_scale} are shown in Table \ref{tab:time_scales_table}.

 Figure~\ref{fig:time_scale} shows the similarity of dropped data blocks across the time scale (Phase 2 shown on the X-axis and Phase 4 shown on the Y-axis) for the 124M (left) and 345M (right) parameter models. We find that there is a higher similarity in the data points dropped by the 124M parameter model in Phases 2 and 4 than in the case of the 345M parameter model (mean, variance, and standard deviation are provided in Table~\ref{tab:time_scales_table}). This behavior signals that the lower capacity of the 124M parameter model inhibits its learning process in Phase 3, such that it finds similar points confusing in Phases 2 and 4. In contrast, the 345M parameter model learns the data blocks it found confusing in Phase 2 by focusing on them, and moves on to learning new complex blocks by Phase 4.  
%  The mean similarity of data blocks dropped byWe find that the 345M model drops more similar data points across time, than the 124M model (Figure~\ref{fig:time_scale}). The mean similarity for the 
% This might be because the larger model retains more information within one epoch than the smaller model, and can discriminate between easy and hard samples from early on.
% Conversely, the smaller model with 124M parameters appears to perceive different data blocks as easy or hard across the training time, indicating that it understood lesser within one epoch.

We conduct a similar study in order to characterize the similarity in data blocks across model sizes. Figure~\ref{fig:model_scale} plots the cosine similarity heatmap for the data blocks dropped by the 124M parameter model (X-axis) and those dropped by the 345M parameter model (Y-axis) in Phase 2. The mean, variance, and standard deviations of the cosine similarity are 0.38, 0.15, and 0.023, respectively. This indicates that the data blocks found easy and dropped in Phase 2 by both models display a moderate level of similarity, but also differ significantly. 
% Therefore, static data selection techniques based on distance metrics \cite{tirumala2023d4, abbas2023semdedup, kaddour2023minipile, xie2023data} which select a fixed subset to train for the entire training duration do not scale well to the diversity in text. Based on our analysis in Figure \ref{fig:model_scale} and \ref{fig:time_scale}, different data blocks are found difficult by models of different capacities at different training instants, driving the need for dynamic data selection methods like LFR.

Finally, we observe the cosine similarity of data blocks dropped and retained during phase 4 for the 124M (left) and 345M (right) parameter models in Figure~\ref{fig:retained_dropped}. The mean, standard deviation, and variance are detailed in Table \ref{tab:retained_droppped}. The smaller model displays greater similarity (lighter values in the heatmap) between the dropped and retained blocks than the larger model. We hypothesize that the larger model can perform reasonably well across similar data points, but struggles with very different complex blocks by the fourth training phase. In contrast, the smaller model does not display the same high-level of understanding (similar perplexity values) on related data blocks. 

To summarize, \textbf{data block importance varies across training time, and across model sizes}. Therefore, static data selection techniques \cite{tirumala2023d4, abbas2023semdedup, kaddour2023minipile, xie2023data} which select a fixed subset to train for the entire training duration for all model architectures do not adapt to the changing training dynamics of LLMs. Based on our analysis in Figure \ref{fig:model_scale} and \ref{fig:time_scale}, different data blocks are found difficult by models of different capacities at different training instants, driving the need for dynamic data selection methods like LFR.
We detail further analysis on the selected and discarded data blocks and demonstrate how models initially focus on learning conversational and anecdotal data, before proceeding to learn factual data in Appendix~\ref{app_ablation_extension}.

\subsection{Extended Analysis on Dropped and Retained Data Blocks for GPT-2}\label{app_ablation_extension}

\begin{table}[]
    \centering
    
    \begin{tabular}{|c|c|c|c|}
    \hline
        Model & Mean & Std & Variance \\
        \hline
        GPT-2 124M & 0.45 & 0.20 & 0.04\\
        GPT-2 345M & 0.30 & 0.12 & 0.01 \\
        \hline
    \end{tabular}
    \caption{Mean, standard deviation (std), and variance of cosine similarity matrices for dropped data blocks across time scale (Phase 2 and Phase 4) for the GPT-2 124M and 345M models.}
    \label{tab:time_scales_table}
\end{table}

\begin{table}[]
    \centering
    
    \begin{tabular}{|c|c|c|c|}
    \hline
        Model & Mean & Std & Variance \\
        \hline
        GPT-2 124M & 0.44 & 0.21 & 0.046\\
        GPT-2 345M & 0.32  & 0.13 &  0.018\\
        \hline
    \end{tabular}
    \caption{Mean, standard deviation (std), and variance of cosine similarity matrices for dropped and retained data blocks in Phase 4 of pretraining for the GPT-2 124M and 345M models.}
    \label{tab:retained_droppped}
\end{table}

\begin{figure}
  \includegraphics[width=\columnwidth]{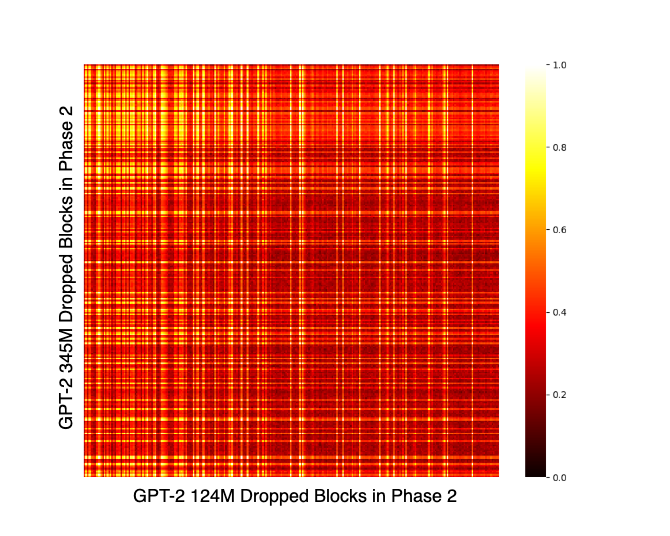}
  \caption{Cosine similarity heatmap for data blocks dropped during Phase 2 of GPT-2 124M and 345M pretraining shows moderate similarity, indicating different data points are considered easy by each model.}
  \label{fig:model_scale}
\end{figure}

\begin{figure*}
  \includegraphics[width=\textwidth]{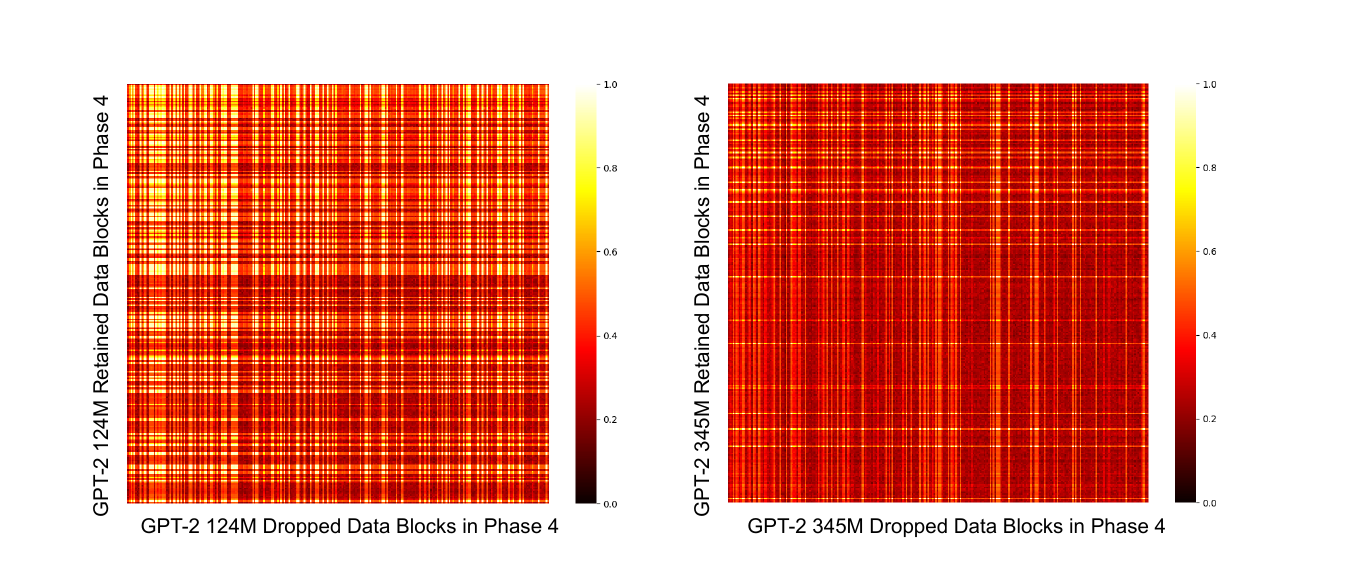}
  \caption{Cosine similarity heatmaps for dropped and retained data blocks during Phase 4 of pretraining for the GPT-2 124M (right) and 345M (left) models.}
  \label{fig:retained_dropped}
\end{figure*}
\begin{figure}
\includegraphics[width=\columnwidth]{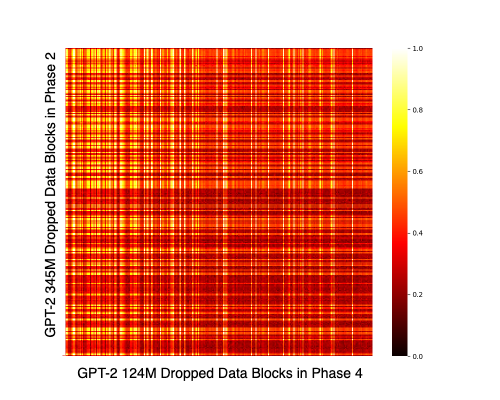}
  \caption{Cosine similarity heatmap for dropped data blocks during Phase 4 of GPT-2 124M and Phase 2 of the 345M model.}
  \label{fig:time_model_mix_1}
\end{figure}

% \begin{figure*}[]
%   \includegraphics[width=\textwidth]{./latex/Figures/dropped_retained_earlier.drawio.pdf}
%   \caption{Cosine similarity heatmaps for dropped and retained data blocks during Phase 2 of pretraining for the GPT-2 124M (right) and 345M (left) models.}
%   \label{fig:retained_dropped_appendix}
% \end{figure*}

\begin{table}[]
    \begin{tabular}{|c|c|c|c|}
    \hline
        Model & Mean & Std & Variance \\
        \hline
        GPT-2 124M & 0.42 & 0.19 & 0.04 \\
        GPT-2 345M &  0.40 & 0.18 & 0.03 \\
        \hline
    \end{tabular}
    \caption{Mean, standard deviation (std), and variance of cosine similarity matrices for dropped and retained data blocks in phase 2 of pretraining for the GPT-2 124M and 345M models.}
    \label{tab:retained_droppped_appendix}
\end{table}

\begin{table}[]
    \centering
    \begin{tabularx}{\columnwidth}{|X|}
    \hline
    Become a fan of Slate on Facebook. Follow us on Twitter.The first time I crocheted a soccer ball was on the occasion of the 2010 World Cup. It was being held on the continent of Africa, and I thought the African Flower hexagon motif was the perfect vehicle for a crochet soccer ball celebrating the continent’s first time hosting the World Cup: This time around, instead of using all 9000 of my favorite colors, I limited myself to the colors of the flags of the thirty-two countries that had made it to the final rounds of the World Cup competition, and I did my best to incorporate the designs of their flags into the thirty-two hexagons and pentagons of a soccer ball.\\
\hline
        ML-77 Missile Launcher: Based on existing technology, the ML-77 is a rapid-fire missile launcher using seeking projectiles. Each projectile features a friend-or-foe recognition system, ensuring it will find a hostile target even if the user's aim is not completely accurate. The locking mechanism of the ML-77 allows the shooter to ignore cover and line of sight when shooting at locked on enemies, though an attack roll is still required. Locking on to an enemy requires a move action when the enemy is in line of sight and lasts for the rest of the encounter, or until a new target is locked. \\
    \hline
    % \hline
    \end{tabularx}
    \caption{Examples of text dropped by the 345M model in phase 2 (top) and phase 4 (bottom). }
    \label{tab:dropped_time}
\end{table}

\begin{table}[h]
    \centering
    \begin{tabularx}{\columnwidth}{|X|}
    \hline
    Unofficial reports claimed the car was powered by a 95kW 1.5-litre non-turbo petrol engine but Tada didn't confirm. When asked what powers the S-FR Tada revealed he was considering three choices. "When you see the S-FR concept I suppose you imagine it is a 1.5-litre car but nowadays I can choose many kind of engines," he explained. "Downsized turbo, 1.5-litre naturally aspirated and something additional as well. Now we are thinking which one is the best engine for a small sports car." Tada also admitted that the company is unlikely to turn to a partner like it did with Subaru for the 86/BRZ or the new 'big brother' sports car with BMW.
\\
\hline
       In April, MYIR released a Linux-powered MYS-6ULX single board computer, which was notable for being available in two different versions using NXP’s low power, Cortex-A7 i.MX6 UltraLite (UL) or the more affordable, and almost identical i.MX6 ULL SoC. Now, MYIR has released an “MYB-6ULX Expansion Board” designed to stack onto either model. The \$21.20 accessory adds a second 10\/100 Ethernet port to the MYS-6ULX, as well as new CAN, RS485, audio, micro-USB, RTC, and camera functions. MYB-6ULX Expansion Board with MYS-6ULX (left) and detail view (click images to enlarge). The MYB-6ULX Expansion Board has the same 70 x 55mm dimensions as the MYS-6ULX, which is available in two models: The i.MX6 UL based MYS-6ULX-IND has -40 to 85°C support instead of 0 to 70°C, and the i.MX6 ULL based MYS-6ULX-IOT features a USB-powered WiFi radio. The 4-layer expansion board runs on 5V power, and shares the industrial temperature support of the IND model. \\
    \hline
    % \hline
    \end{tabularx}
    \caption{Examples of text retained by the 345M model in Phase 2 (top) and Phase 4 (bottom). }
    \label{tab:dropped_time_retained}
\end{table}

\begin{table}[h]
    \centering
    \begin{tabularx}{\columnwidth}{|X|}
    \hline
    In the book, the mythical California is ruled by Queen Califa and populated only with female warriors who brandish gold weapons. They even harness their animals in gold because it is the only mineral on the island. The legend of Califa and her island was well known among New World explorers. In 1536 when Hernán Cortéz arrived in Baja California, he believed he had landed on the legendary island. Over three hundred years later gold was discovered in California, making the legend partially true and earning the state its nickname: The Golden State. \\

    % \hline 
    % One commenter on Guyenet's piece points out, "It will asymptote towards 100\% but never get there, as there is always some protein and fat," but that disclaimer doesn't make for sexy Health Department ad copy. What might work? "The Future is The Fly."Glasses broke and furniture moved at Premier House, the official prime ministerial residence in Wellington, Mr. Key said. He had stayed there overnight after meeting with Secretary of State John Kerry, who was on a state visit there and to Antarctica.\\
\hline
        Segregated Witness, defined by Bitcoin Improvement Proposal 141 (BIP141), was deployed using an activation mechanism (BIP9) that requires 95 percent of all miners (by hash power) to signal support for the upgrade within the span of a two-week difficulty period. That’s at least 1916 blocks within 2016 blocks, to be exact. This threshold has just been reached. While the current difficulty period will not end until tomorrow, all blocks in this difficulty period are signaling support for the upgrade so far. This now totals over 1916 of them. \\
    \hline
    % \hline
    \end{tabularx}
    \caption{Examples of text dropped by the 124M model in Phase 2 (top) and Phase 4 (bottom). }
    \label{tab:dropped_time_small}
\end{table}

\begin{table}[h]
    \centering
    \begin{tabularx}{\columnwidth}{|X|}
    \hline
     to the GUI installer. Most notably there is no support for configuring partition layout, storage methods or package selection. Please refer to the official documentation for details. Here you can find some useful information on creating and using kickstart files which can be used to perform advanced configuring without the need for the GUI installer. The message "Insufficient memory to configure kdump!" appears during install. This is a known issue which appears on systems with less than 2 GB RAM. This can be ignored. Content for both the i386 and x86\_64 architectures is split into two DVDs. We have tried to get all basic server and basic desktop installs only from DVD-1. Make sure that you setup correctly the selinux context of the public key if you transfer it to a CentOS 6 server with selinux enabled. 
\\
\hline
    Once you signed up, you can either click on the Todo tab or the sign in link to gain access to the application. Note that if you are selecting sign in in the same session in which you signed up, you will automatically sign in with the same account you used for signing up. If you are signing in during a new session, you will be presented with Azure AD's credentials prompt: sign in using an account compatible with the sign up option you chose earlier (the exact same account if you used user consent, any user form the same tenant if you used admin consent). If you try to sign-in before the tenant administrator has provisioned the app in the tenant using the Sign up link above, you will see the following error.

    % \hline 
    % One commenter on Guyenet's piece points out, "It will asymptote towards 100\% but never get there, as there is always some protein and fat," but that disclaimer doesn't make for sexy Health Department ad copy. What might work? "The Future is The Fly."Glasses broke and furniture moved at Premier House, the official prime ministerial residence in Wellington, Mr. Key said. He had stayed there overnight after meeting with Secretary of State John Kerry, who was on a state visit there and to Antarctica.\\
         \\
    \hline
    % \hline
    \end{tabularx}
    \caption{Examples of text retained by the 124M model in phase 2 (top) and phase 4 (bottom). }
    \label{tab:dropped_time_small_retained}
\end{table}

In this section, we expand on the ablation study in Section \ref{ablation-dropped-retained} in order to better characterize the data blocks considered easy / hard. 

Tables~\ref{tab:dropped_time} and \ref{tab:dropped_time_small} provides examples of text blocks dropped in Phases 2 and 4 by the 345M and 124M parameter models respectively. Similarly, Tables~\ref{tab:dropped_time_retained} and \ref{tab:dropped_time_small_retained} provide examples of data blocks retained by the models in Phases 2 and 4. We printed out and went over all the text dropped and retained in both Phases, and notice that text considered easy in phase 2 was more conversational, and those considered easy in phase 4 were more factual. This might indicate that the model first learned conversations and personal anecdotes, before being able to retain factual information. These findings are further corroborated by the examples of data retained in both phases. We are working on further analysis across different model families and sizes to strengthen this understanding.

Next, we continue the analysis of the cosine similarity heatmaps evaluated across training time and model parameter scales presented in Section~\ref{ablation-dropped-retained}. Here, we answer the following questions:
\begin{enumerate}
    \item Are there similarities in the data blocks considered easy and dropped in Phase 4 of training of the 124M parameter model with those considered easy and dropped by the 345M parameter model in Phase 2? 
    \item Are the data blocks dropped similar to those retained at any given training phase? Note that Section~\ref{ablation-dropped-retained} presented this analysis only for Phase 4 of the 124M and 345M parameter models in Figure~\ref{fig:retained_dropped}.
\end{enumerate}

Figure~\ref{fig:time_model_mix_1} depicts the cosine similarity heatmap for the data blocks dropped by the 124M parameter model in Phase 4 (X-axis) with those dropped by the 345M parameter model in Phase 2 (Y-axis). The mean, standard deviation, and variance of the similarity are 0.43, 0.18, and 0.03 respectively. In contrast, the mean cosine similarity of data blocks dropped in Phase 2 of pretraining of both the models was 0.38 (Section~\ref{ablation-dropped-retained} and Figure~\ref{fig:model_scale}). This indicates that the smaller model "catches up" with the knowledge accumulated by the larger model, and considers similar block easy in Phase 4 as those considered easy by the larger model in Phase 2. 

% Next, in Figure~\ref{fig:retained_dropped_appendix}, 
Next, we plot the cosine similarity heatmap for the dropped and retained data blocks in Phase 2 for the 124M (left) and 345M (right) parameter models. The mean, variance, and standard deviations of the similarity are shown in Table~\ref{tab:retained_droppped_appendix}. Observing the mean similarity value and heatmap in Table~\ref{tab:retained_droppped} and Figure~\ref{fig:retained_dropped}, we find that the cosine similarity for dropped and retained data blocks is higher in Phase 2 than Phase 4 in case of the 345M parameter model. In contrast, the value remains high in both Phases for the 124M parameter model. This finding indicates that both the smaller and larger model start the training by being confused about similar data blocks. However, the larger capacity of the 345M parameter model allows it to learn the dataset well in Phases 2 and 3, and move on to more complex data blocks in Phase 4 (thus reducing the mean similarity in Phase 4). The smaller model continues remaining unsure about similar data blocks. Since we observed that the smaller model "catches up" with the training of the larger model (in Figure~\ref{fig:time_model_mix_1}), we hypothesize that the smaller model will eventually display similar behaviour as the larger model once trained for longer iterations.

\end{document}